\documentclass{arxiv}

\usepackage{graphicx,amsmath,bm}

\usepackage{setspace}

\usepackage{epstopdf}
\usepackage{color}
\usepackage{tablefootnote}

\usepackage[hyphens]{url}

\usepackage{float}
\usepackage{subcaption}
\usepackage{listings}
\usepackage{mathalfa}
\usepackage{gensymb}
\usepackage{appendix}
\usepackage{hyperref}
\usepackage{amssymb}
\usepackage{multicol}

\newcommand\abovePi{\mathrel{\overset{\makebox[0pt]{\mbox{\normalfont\tiny\sffamily T}}}{\prod}}}

\begin{document}

\title{Monocular Imaging-based Autonomous Tracking for Low-cost Quad-rotor Design - TraQuad}


\author{Lakshmi Shrinivasan\textsuperscript{1,*}\and Prasad N R\textsuperscript{2}}
\affilOne{\textsuperscript{1} Assistant Professor, Electronics and Communication,  M S Ramaiah Institute of Technology\\}
\affilTwo{\textsuperscript{2} Alumnus, Electronics and Communication,  M S Ramaiah Institute of Technology}


\twocolumn[{

\maketitle

\begin{abstract}
TraQuad is an autonomous tracking quadcopter
capable of tracking any moving (or static) object like cars, humans, other drones or any other object on-the-go. This article describes the applications and advantages of TraQuad and the reduction in cost (to about 250\$) that has been
achieved so far using the hardware and software capabilities and our custom algorithms wherever needed. This description is backed by strong data and the research analyses which have been drawn out of extant
information or conducted on own when necessary. This also describes the development of completely autonomous (even
GPS is optional) low-cost drone which can act as a major platform for further developments
in automation, transportation, reconnaissance and more. We describe our ROS Gazebo simulator and our STATUS algorithms which form the core of our development of our object tracking drone for generic purposes.
\end{abstract}

\msinfo{}{}{}


}]


\setcounter{page}{1}
\corres
\volnum{1}
\issuenum{1}
\monthyear{January 2018}
\pgfirst{1}
\pglast{100}
\doinum{1}
\articleType{}

Note:\footnote{This article was drafted on January 6 2017 and was revised on 21 January 2018}


\markboth{Lakshmi et al}{Monocular imaging and control of TraQuad}

\section{Introduction}
The very term 'quad-rotor' refers to the simple design of an aerial system consisting of four propellers which does not necessarily involve overall system's movement as in planes or dynamically adjust the propeller angles as in helicopters. This design is quite feasible for low-cost exploration and reconnaissance designs having VTOL built-in and thus will probably replace 'private planes' and cars and aid in space-missions where reusable-rockets with relatively dense atmosphere can prove to be costly. This proved to be the inspiration for the development of autonomous navigation to enable the usage of capable drones by consumers as a whole. The follow-me mode is being used in adventure sports to explorations in hazardous environments. We describe a drone system that can be more advanced and discuss the scientific and practical implications.
\\\\
This article describes the novelty in two aspects:

a. The performance and the ease of development of our software needed for tracking according to the requirements of the hardware. (Our STATUS algorithm is reliable even on memory constrained Android phones)

b. The performance of our algorithms in power constrained devices (like Android phones which was used for on-field tests) and embedded systems (validated by simulations).

\section{Related works}

We attempted to build follow-anything drone and we found that there were many libraries and softwares which are capable of executing more complicated tasks than what we are doing. But, we realised that there is no sufficient algorithm meant for drone which uses both physical dynamics and software for manoeuvres. Several survery papers on object tracking gave us lot of insights about the tracking procedures used and some algoritms like TLD even use learning mechanisms\cite{2006_24}\cite{2011_23}\cite{2014_25}. But, they only use image-processing based software-only approach. Moreover, we found that some algorithms were extraordinary when used in combination (for example tuned Kalman Filter with frequency domain operations); But, they worked only in specific conditions as mentioned by the parameters. Thus, we decided to build an image-processing drone ourselves and test it both on hardwares and in ROS Gazebo software in the loop simulations. We have relatively recently found that there is a simulator built using Unreal Engine\cite{45degreeSimulator}. This paper effectively consolidates most of the object tracking drone's efforts. But, threading is not well supported even in binaries in Unreal Engine and thus, we continued our decision to build a physics based software simulator in ROS Gazebo using ODE. Stanford Driving Software helped us a lot in understanding the optimal use of Gaussians in tracking roads and the use of ROS (some of the algorithms which we used while devising and deriving STATUS)\cite{stanley34}. The resultant Gaussian that we derived using first principles of probability can be found in \hyperref[Appendix C]{Appendix C}. Falkor systems had designed a similar object tracking drone using just selecting feature-matching that is hard-coded for their logo \ref{Appendix 12}. Although their company has shut down, their open-source software gave us an insight about the actual performance optimisation for object tracking drone and they have done it really well although it does not track even the logo given the specific tuning of PID parameters. Thus, we were left with KF, EKF, TLD, DCT + DFT, FFT or a combination of all with ensemble learning. We found training of that difficult and reduced the algorithm to STATUS and this has been described with both hardware and software.

\section{Choice of embedded system based quadrotor design}
\label{initial_design}

The design choices were made not by examining the standard components or the approximate cost that conventional wisdom suggests. Instead, the costs were evaluated mainly according to the simple laws of physics. Much of the drone's cost would depend on it's flight; Thus a general assumption was made about the density of air and wind speed as follows:
\\\\
a. 1.225 Kg/$m^3$ is the sea-level air density and the quadcopter can only fly upto 400 feet as per FAA rules\cite{faa1}. Thus, the air-density at 400 feet (122m) would be 1.2086 Kg/$m^3$. Thus, it is reasonable to assume that air-density would not change by more than 1.34\% as 100(1-$\exp(-122/9042)$)=1.34\%.\cite{flightDynamics2}
\\\\
b. Wind-speed can be between 0-100m/s practically.\cite{flightDynamics2}
\\\\
Thus, a drone practically consumes power only when it flies and wind can be an undesirable force sometimes. Hence, the cost of the drone will be close to zero if the mass of the quadcopter is relatively light-weight; The only object this drone had to support during testing was the processing and sensor unit. Hence, the entire design was considered from the weight of processor-sensor unit with a relatively huge margin to accommodate for intensive testing. Beaglebone Black was chosen as the processor initially, ultrasonic sensor was chosen for critical obstacle avoidance and critical sensor data fusion. ArduPilot APM 2.6 was chosen for flight control. A 2200 mAh Lithium polymer battery was installed, four 850KV (850 rpm/volt) brushless motors, four integrated BEC-ESCs, connectors, four carbon-nylon propellers and related experimental setups' components were purchased. Camera and WiFi module were thought of in terms of the bandwidth of USB and Android phone and will be described in detail. A 3D design of the chassis was designed; However, local chassis was sufficient for quicker prototyping although it lacked customisation. We have never bought expensive drone RC transmitter available in stores.

\subsection{Choice of components}
\label{choice_of_components}

These hardwares were chosen as per the requirement after thorough research without much consideration for preconceived notions like thumb-rules. Comparison of embedded processors has been described in \hyperref[Table 1]{Table 1}. Comparison of Flight controllers has been described in \hyperref[Table 2]{Table 2}.

\subsubsection{Processor}
Beaglebone Black has 512 MB DDR3 RAM, 4 GB eMMc with external micro-SD card as an option, 1GHz ARM Cortex A8 processor; It even features 2 PRU units which helps in real-time control of the vehicle and the software for PRUs is in assembly and C/C++ languages while the main-line operating system supported is Angstrom (Debian variant), supports NEON acceleration and it supports many other important OSes as well. It features SGX530 graphical processor and U-boot for better management of hardware-software booting system which we thought of as optional features (\ref{Appendix 1}). Beaglebone Black is the only standard board which has 3 independent processors which can run 2 different type of platforms and has official support for an RTOS (Starterware). Beaglebone has an integrated system for PRUs and ARM Cortex processor which means that both real-time software and high-level software languages work synchronously. The power consumed is about 10 watts at peak while the board itself weighs about 40 grams.\cite{BBBdatasheet3} Debian operating systems are POSIX compliant and offer a lot of standard programming languages' software tools like that of C/C++, FORTRAN, Perl, Python, PHP, JavaScript and options to install additional SDKs.\cite{debian4} This was the best board available to us for an integrated system like drone which requires real-time pin control and image-processing and inter-frame image processing.
\\\\
Without Beaglebone, the only standard method to achieve both non-real-time and real-time requirements was to use two microcontrollers with time-stamped data-transfer. Otherwise, another option was to develop separate Linux kernels on different cores if the boot-sequence of BIOS/UEFI permitted so.\cite{twinLinux5}\cite{jailhousing} We were also considering developing our own operating system if needed. (It wasn't on Yocto Linux project; it was on SUSE linux titled "Prasad N R's linux distro" (simple-drag-and-drop is nice in Suse Studio) which was working great for software-programming. \cite{suse20})\\

Adapteva's Parallella board which is meant for low-power parallel computations was also considered; However, that suffers from the problem of low-power FPGA devices (\ref{Appendix 20}a) (but provides 2 micro-USBs (\ref{Appendix 20}b)). FPGA with run-time re-configurability of logic-blocks with software-tools like myHDL (\ref{Appendix 20}f) was considered; However, a research on the performance has concluded that FPGA can generally be used for low-power devices while a combination of CPU and GPU can mean extraordinary performance (\ref{Appendix 20}e). We were left with two options Olimex's AM3352-SOM series of boards and Beaglebone Black. As AM3359x is manufactured by Texas Instruments officially and Beaglebone is a standard-board having a strong community-support, we chose Beaglebone Black (\ref{Appendix 20}c) (\ref{Appendix 20}d). (We came across BeagleBone Blue which has low-power profile. We also came across Seeed Studio's BeagleBone Green which has support for lot of sensors through Grove connectors. We intend to research upon this. But, we believe that low-power BeagleBone and restricting ourselves to SeeedStudio's components might hinder the our custom high performance image-processing algorithms. BeagleBone Black has profile of application and not Microcontroller which has low-power mode.(\ref{Appendix 30}))

\begin{figure}[h]
\label{Figure 1}
\centering
\includegraphics[width=\linewidth]{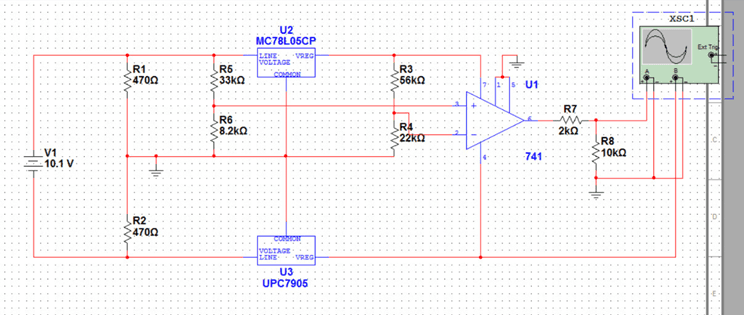}
\caption{Screen-shot of our voltage-level circuit-design designed to indicate LiPo battery level to BeagleBone}
\end{figure}

Beaglebone had to be powered using power-bank as the other option was to use external custom circuit as in \hyperref[Figure 1]{Figure 1}. We found that analog interrupt (not comparator interrupt) can be noisy if the quantization of the Beaglebone's ADC is lesser than the resolution of the input.

Note for Table 1\footnote{Note: As of 21 January 2018, we have found Intel EUCLID and UP boards both of which seem to be proprietary. Also, the website https://click.intel.com/intelr-euclidtm-development-kit.html redirects to 404 Error Page Not Found. Shipping of up-boards for backers on Kickstarter seems to have started. But, we have not backed UP boards.(\ref{Appendix 29})}

\begin{table*}[ht]
\label{Table 1}
  \centering
  \resizebox{\textwidth}{!}{
    \begin{tabular}{|c|c|c|c|c|c|}
    \hline
    \textbf{Name of the board} & \textbf{Processor} & \textbf{Website} & \textbf{Operating Systems supported} & \textbf{RAM} & \textbf{Max clock speed} \\ \hline
    Beaglebone Black & ARM Cortex A8 & \textcolor[rgb]{ .02,  .388,  .757}{www.beagleboard.org/black} & \shortstack{Angstrom, Ubuntu, \\Android and RTOS Starterware} & 512MB & 1GHz \\ \hline
    UDOO  & ARM Cortex A9 & \textcolor[rgb]{ .02,  .388,  .757}{www.udoo.org} & Ubuntu and Android variants & 1GB   & 400-528 MHz \\ \hline
    Arduino & ATMEGA & \textcolor[rgb]{ .02,  .388,  .757}{www.arduino.cc} & RTOS/none & 2KB-8KB & 16MHz \\ \hline
    Odroid & Amlogic S805 & \textcolor[rgb]{ .02,  .388,  .757}{www.hardkernel.com} & \shortstack{Ubuntu and Android\\ variants with OpenELEC} & 2GB   & 1.5GHz \\ \hline
    Raspberry Pi & Broadcom BCM SoC, ARM Cortex A7 & \textcolor[rgb]{ .02,  .388,  .757}{www.raspberrypi.org} & \shortstack{NOOBS and Raspian\\ (3rd party RTOS is\\not supported officially)} & 1GB   & 1.2GHz \\ \hline
    Pcduino & AllWinner A10 SoC, ARM Cortex A8 & \textcolor[rgb]{ 0,  .439,  .753}{www.pcduino.com} & Ubuntu and Android & 1GB   & 1GHz \\ \hline
    Pandaboard & ARM Cortex A9 & \textcolor[rgb]{ .02,  .388,  .757}{www.pandaboard.org} & Minix3, Ubuntu, Android and Angstrom & 1GB   & 1.2GHz \\ \hline
    Minnow Board & Intel Atom E640 & \textcolor[rgb]{ .02,  .388,  .757}{www.minnowboard.org} & Yocto, Wind River, Android, Windows & 1GB   & 1GHz \\ \hline
    Gumstix & \shortstack{Many proprietary versions are available \\with limited customisations (Geppetto)} & \textcolor[rgb]{ .02,  .388,  .757}{www.gumstix.com} & Linux generally & \shortstack{Depends on\\customisation} & 1Ghz generally \\ \hline
    Teensy & ATMEGA to 32 bit ARM Cortex M4 & \textcolor[rgb]{ .02,  .388,  .757}{www.pjrc.com/teensy/index.html} & None  & 2KB-64KB & 16-64 MHz \\ \hline
    Radxa (Rock) & ARM Cortex A9 & wiki.radxa.com & Debian and Android & 2GB   & 1.6GHz \\ \hline
    Olimex (AM3352-SOM series) & ARM Cortex A8 & \textcolor[rgb]{ .02,  .388,  .757}{www.olimex.com} & \shortstack{Angstrom, Ubuntu, Android\\and RTOS Starterware} & 512MB & 1GHz \\ \hline
    CubieBoard & AllWinner A80 SoC, ARM Cortex A15 & \textcolor[rgb]{ .02,  .388,  .757}{www.cubieboard.org} & Linux and Android & 2GB   & 1.3GHz \\ \hline
    Phidgets SBC3 & FreeScale i.MX28 & \textcolor[rgb]{ .02,  .388,  .757}{www.phidgets.com} & Debian Linux & 128MB & 454MHz \\ \hline
    Intel Galileo (Retd. Product) & Intel Quark SoC X1000 & \textcolor[rgb]{ .02,  .388,  .757}{\shortstack{www.arduino.cc/en/Arduino\\Certified/IntelGalileo}} & None  & 512KB & 400MHz \\ \hline
    STM32 & ARM Cortex M series & \textcolor[rgb]{ .02,  .388,  .757}{www.st.com} & None  & 512KB & 216MHz \\ \hline
    PIC32 & PIC   & \textcolor[rgb]{ .02,  .388,  .757}{www.microchip.com} & None  & 512KB & 252MHz \\ \hline
    Intel Edison & Intel Atom & \textcolor[rgb]{ .02,  .388,  .757}{\shortstack{www.arduino.cc/en/Arduino\\Certified/IntelEdison}} & Linux and RTOS & 1GB   & 500MHz \\ \hline
    NVIDIA Jetson (TK1 and TX1) & ARM Cortex A15-A57 (Tegra SoC) & \textcolor[rgb]{ .02,  .388,  .757}{\shortstack{www.nvidia.com/object/\\embedded-systems-dev-kits-modules.html}} & Linux (Ubuntu) & 2GB-4GB & 2.3GHz \\ \hline
    Ornage Pi & ARM Cortex A7 (AllWinner SoC) & \textcolor[rgb]{ .02,  .388,  .757}{\shortstack{www.orangepi.org/orangepizero/}} & Debian and Android & 256/512MB & 1.2GHz \\ \hline
    \end{tabular}%
    }
    \caption{A brief comparison among standard embedded processors. (Excludes Banana-pi, CHIP and such pre-order and generic non-PRU (lacking multiple independent controllers) and non-standard options)}
  \label{table_boards}%
\end{table*}%

\subsubsection{Ultrasonic sensor}
This is being used for obstacle avoidance as the range is relatively little (about 4 metres of accuracy) and critical data-fusion for relatively near-ranged tracking. This is being used primarily for two reasons: It's weight is virtually zero (unlike many LIDARs and Laser Range Scanners) and the cost is low as well. For relatively long-ranged tracking, Laser sensors would be great. For our experiment, we restricted the drone from moving further if an object was determined to be within a metre approximately. The time-limit mentioned in the datasheet is 60 milli-seconds for a cycle of measurement.\cite{ultrasonic6} This can mean a displacement of 0.3 m if the drone's velocity is 5 m/s or 1.5 m if the drone's velocity is 25 m/s which can be dangerous if the ultrasonic code does not work properly. We had designed real-time GPIO controls for the purpose of MAVlink, ultrasonics and other allied aspects.

\subsubsection{Chassis, battery and remote control}

\begin{figure}
\label{Figure 2}
  \centering
  \includegraphics[width=0.75\linewidth]{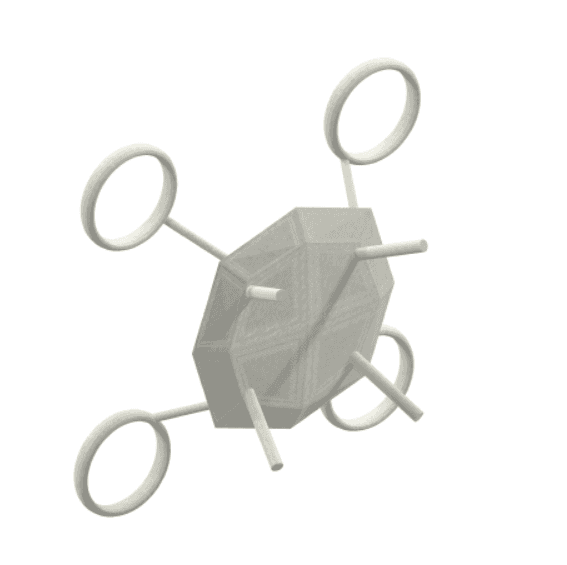}
  \caption{3D design of the custom chassis}
  \label{custom_chassis}
\end{figure}

A custom chassis was designed initially (a primitive design as in \hyperref[Figure 2]{Figure 2} was designed using open-source software Blender (\ref{Appendix 7}) after which a plastic chassis was bought\footnote{There were practical problems of printing in India or bearing the cost of importation}.) Typical expensive RC remote control was eliminated by using Android phone as the controller (we had built custom application for that). Battery was one of the major issues; Generally, more number of batteries mean greater flight time. However, this increase in flight-time is greatly mitigated by the increase in weight because of batteries. The flight time achieved was about 12 minutes for a design-capacity of 3 Kg out of which, we had reduced the entire weight to 1.2 Kg (in the final design). This design was estimated using eCalc website which proved that our quadcopter won't lose control easily for breeze and only loses some control with winds of about 8 m/s speed (\ref{Appendix 9}). Some of the methods commonly used to increase flight-time are dangerous; Usage of fuel-cells and fuel-engines have not resulted in drastic increase of flight-time for drones with relatively heavy-weights and some are dangerous too. Usage of helium-balloons or helium cabins can cause unneeded floatation and reverse-thrust might be necessary (or it might get out of control even then or might provide little thrust of 3 Kg/$m^3$). When we tried to assess some of the unconventional methods, none of the economically feasible solutions emerged. Solar panels for quadcopters is still questionable as multi-junction solar-cells are not economically viable yet although the costs are exponentially decreasing. Perovskite material itself has an efficiency of about 20\% (and contains toxic lead) and can produce about 200W/$m^2$ which can hardly be sufficient sometimes.\cite{solar7} Ionic thrusters are not economically viable yet and so is the case with photonic thrusters and Laser-propulsion systems. Nuclear-kits are not relatively small sized and the critical mass can occupy relatively larger volumes (albeit catastrophes can't be ignored). Thus, solar systems entice great hopes and it is infeasible as of now.

\begin{table*}
\label{Table 2}
\centering
  \resizebox{\textwidth}{!}{
    \begin{tabular}{|c|c|c|}
    \hline
    \textbf{Flight Controller} & \textbf{Software support} & \textbf{Manufacturer (company)} \\ \hline
    APM 2.6/PixHawk/PX4 & Custom board, other standard embedded systems & 3D Robotics \\
    Erle Brain & Custom Board, "Provides support for de-facto standard platforms APM and PX4" & Erle Robotics \\
    KK board & Proprietary & HobbyKing \\
    OpenPilot & Discontinued (Only STM32 was supported) & None \\
    MultiWii & Software for Arduino & None \\
    AeroQuad & Software for Arduino (limited sensor support) & AeroQuad \\
    Navio & Mainly for Raspberry Pi with real-time kernel & Emlid \\
    HobbyKing Pilot Mega & Proprietary & HobbyKing \\
    SLUGS & MATLAB and Simulink based software & University of California Santa Cruz \\
    Paparazzi & Softwares for STM32 and LPC2100, Ubuntu & Not officially maintained by any organisation\\
    DJI A2, NAZA-M V2 and WooKong M & Proprietary and only some of the SDKs are open-source softwares & DJI \\ 
    VRbrain & (Unfunded and Closed on Indiegogo) Support for STM32 mainly & VirtualRobotix\\
    Intel Aero-compute & Softwares of Drotek, Ardupilot, PX4, Airmap, Dronesmith, FlytBase (Custom Yocto Linux) & Intel\\
    \hline
    \end{tabular}
    }
    \begin{tablenotes}
          \tiny
          AeroQuad's further info: \url{http://aeroquad.com/showwiki.php?title=Recommended-components-for-your-AeroQuad} \hfill
          Paparazzi's further info: Details of project: \url{https://wiki.paparazziuav.org/wiki/Paparazzi\_vs\_X}
    \end{tablenotes}
    \caption{Comparison table for various standard flight-controllers}
  \label{flight_controllers}
\end{table*}

\subsubsection{Flight Controller}

APM 2.6 was one of the best choices and it is open-source which meant that we can be able to upload our own firmware if needed (\ref{Appendix 2}a). It supports PWM, USB communication and MAVlink commands via telemetry port. The peak power consumed is about 2.5W (\ref{Appendix 2}b). It has an internal barometer (\ref{Appendix 2}c), gyroscope (\ref{Appendix 2}d)(\ref{Appendix 2}e), compass (\ref{Appendix 2}f) accelerometer (\ref{Appendix 2}g) and supports relatively large number of external sensors (including some of Laser sensors and Analog inputs) (\ref{Appendix 2}h). Although APM 2.6 board does not support software-threading by default (\ref{Appendix 2}i), it has 16MB on-board memory (\ref{Appendix 2}j), supports a plethora of modes including battery fail-safe (\ref{Appendix 2}k) while supporting a decent acceleration of about 2.5 m/$s^2$, velocity of about 5 m/s (loiter mode) to 10 m/s (stabilize mode), (\ref{Appendix 2}l) supports upto 18V and 90A ESCs (\ref{Appendix 2}m) and two-way-communication ESCs (UAVCAN) as well (\ref{Appendix 2}n). While we found softwares for those drones like that of Parrot, we didn't find separate ready-made flight-controllers. We found the guides and software-codes of Dronecode useful; As APM and PX4 of 3Drobotics started Dronecode organization, we didn't like the idea of risking our money on proprietary controllers of non-Dronecode companies like those of DJI (\ref{Appendix 10}a)(\ref{Appendix 10}b). This list also excludes those which are quite costly and are invite-only and untested type as with Percepto (\ref{Appendix 13}). 3DR was manufacturing hardwares also when we were performing our research and this was one of the compelling reasons as other companies Erle Robotics were using their custom version of ArduPilot with major developments commited to master branch (\ref{Appendix 25}). (Note: 3DR has shut-down hardware manufacturing unit as of October 5 2016)

\begin{figure}
\label{Figure 3}
  \centering
  \includegraphics[width=0.75\linewidth]{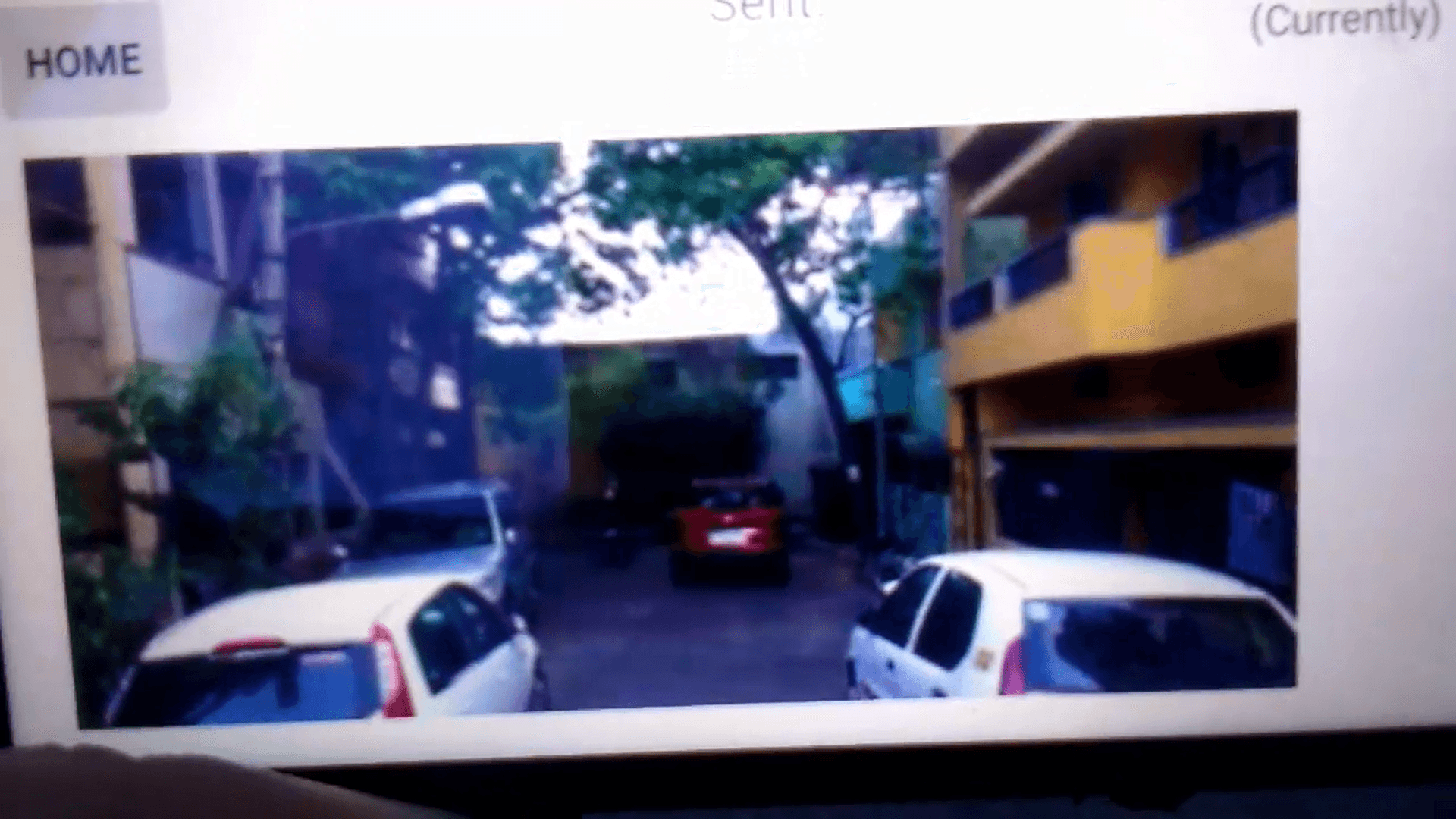}
  \caption{Live-streaming display in autonomous-mode in Android controller phone}
  \label{live_stream_Android}
\end{figure}

\subsubsection{Camera and WiFi}

We tried to interface a low-cost generic USB webcam. However, that did not work well because of two reasons: Many vendor support their webcams only on Windows. Plug-and-play webcams generally transfer only uncompressed data (compressed data is only for Windows in such systems) and this can mean USB bandwidth limitations in Linux Ubuntu or Linux Angstrom. UVC driver is generally linked with main-line kernel (as with all 'drivers' of Ubuntu) and cameras with built-in MJPEG/H264 compression formats are pretty costly. Beaglebone’s camera capes: open-source 3.1MP Camera Cape (\ref{Appendix 3}a) and HD Camera Cape (\ref{Appendix 3}b) were the only standard options (although they expect some of the GPIOs and that gets worse with OS in eMMc) left except the generation of custom board designs. For machine-learning based image-processing algorithms, we have developed an algorithm STATUS (Simultaneous Training And Testing Using Statistical-gaussians) which primarily involves inter-frame image processing and requires real-time computationally light-weight machine learning. PixyCam (\ref{Appendix 4}a) and OpenMV (\ref{Appendix 4}b) turned out to be other options for consideration. However, PixyCam has a button that is used for training. Automation of that training part required significant changes in the design of the camera (using open-source repository and is non-standard method). OpenMV does not provide nice frame-rate (160$\times$120 @ 20fps not 640$\times$480 @ 30fps) and it was no better than the frame rates that we were able to obtain by ourselves (320$\times$240 @ 15fps; Beyond that, we were getting query and argument errors). The OV2640 image-sensor’s datasheet is of preliminary edition (\ref{Appendix 4}c). Recently, JeVois Open-source camera has appeared in the market. But, that does not include Wifi built-in thus making the transmission of video back to user require an intermediate micontroller transmit videos.(\ref{Appendix 31}) Thus, we had to reject those and consider BeagleBone’s official camera modules.
\\\\
WiFi had it’s own issues though. For peer-to-peer WiFi data transfer, there are three methods: Using Ad-hoc networks (But, it is insecure; Thus, Android phone requires software-rooting which we did not like), WiFi-Direct and WiFi-hotspot with Server-client connection. WiFi-Direct devices need certification from WiFi-alliance and WiFi-Direct connection is not guaranteed with legacy WiFi devices (non-WiFi-Direct certified devices).\cite{wifiDirect9} Thus, different versions of Ubuntu Kernel had to be compiled and built to solve WiFi-issues\cite{ubuntuWifi8}. Legacy WiFi Direct (for soft-Access Point) was not working as expected. Thus, we were supposed to choose WiFi-Direct certified device for which we considered the official open-source Beaglebone Black Cape (which supports WiFi-Direct) (\ref{Appendix 5}) owing to problems with legacy WiFi devices. We chose WiFi instead of Bluetooth because of it’s bandwidth and the range (656 feet which increases to about a kilometer with general low-cost range extender with a bandwidth of about 250 Mbps). (\ref{Appendix 6}) We never chose any other unpopular device for consideration. (Other exotic methods like gravitational signaling system or quantum entanglement apparently do not exceed the information-velocity limit (not energy velocity-limit) of assumed space-time and that was not even remotely close to practical application) This was one of the important decisions; Because, Bluetooth is generally easier to program with and consumes virtually zero power while WiFi has an option of extraordinary range and bandwidth.

\begin{figure}[h]
\label{Figure 4}
\centering
\includegraphics[width=\linewidth]{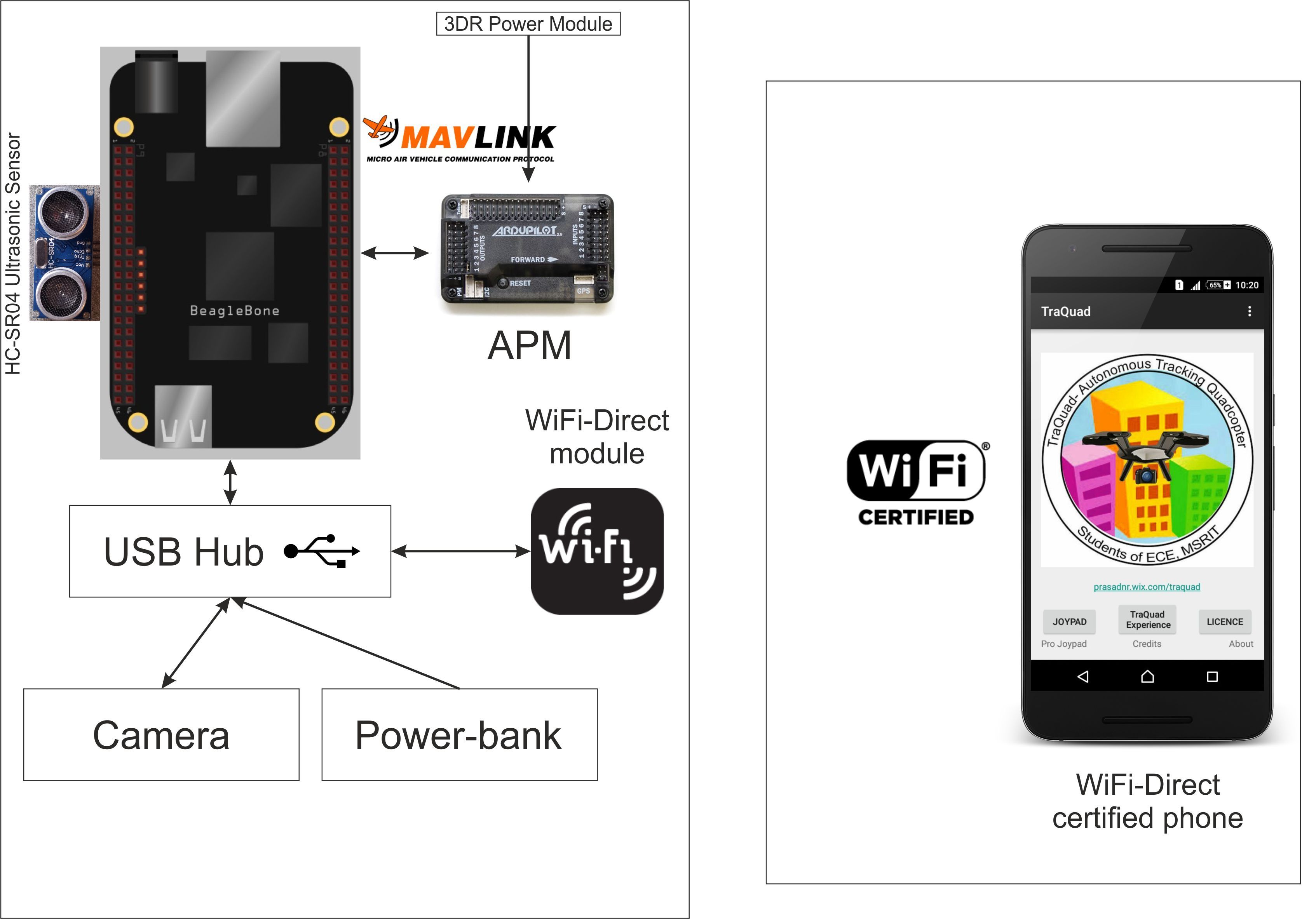}
\tiny Phone has been designed using Device Art Generator \url{developer.android.com/distribute/tools/promote/device-art.html}
\caption{Embedded system based quad-rotor's architecture}
\end{figure}

\subsubsection{Choice of controller}

We chose phone as the controller as laptops are not suited for on-the-go operations of drone. The architecture decided has been shown in \hyperref[Figure 4]{Figure 4}. We chose to have a hand-held system that would cost virtually zero amount. Building a custom hardware for that meant additional costs and we chose to build a software for officially supported Android phone’s unrooted operating system software as an application. We chose Android OS for phone because, only four predominant operating systems namely Android, iOS, Windows and Blackberry exist and only Android is open-source amongst those. Android has the majority of market-share (80.7\%) among the phones.\cite{androidMarket10} We were using API 15 (because we were using VideoView, WebView and other UI elements as in \hyperref[Figure 3]{Figure 3} for controller which didn’t require 3rd party applications excepting WiFi-Direct which was compatible with API 14 or more (\ref{Appendix 6}c)) and developed app using official Android Studio software. Developing app for API 14 phones meant about 98\% percentage popularity for API level and 86.7\% of popularity for normal-sized phone-hardware (and thus great community support) \cite{androidMarket17}.

\subsection{Software design}
This section describes the softare design that was performed for the embedded system and the devices reliant on it.

\subsubsection{Native C/C++ code for real-time control and GPIO}
\begin{figure}[h]
\label{Figure 5}
\centering
\begin{subfigure}{.5\linewidth}
  \centering
  \includegraphics[width=0.95\linewidth]{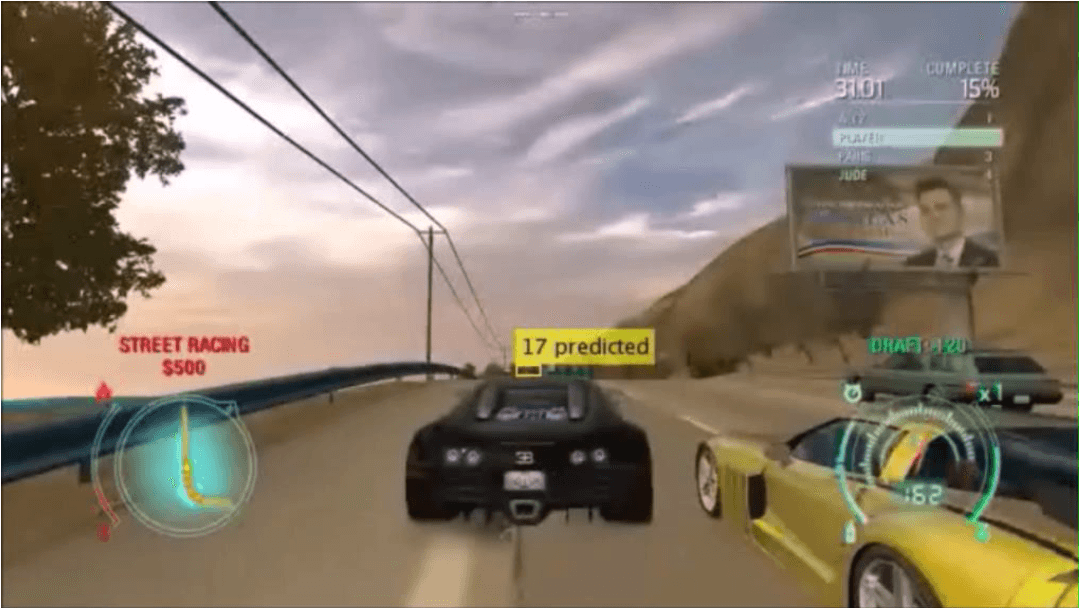}
\end{subfigure}%
\begin{subfigure}{.5\linewidth}
  \centering
  \includegraphics[width=0.95\linewidth]{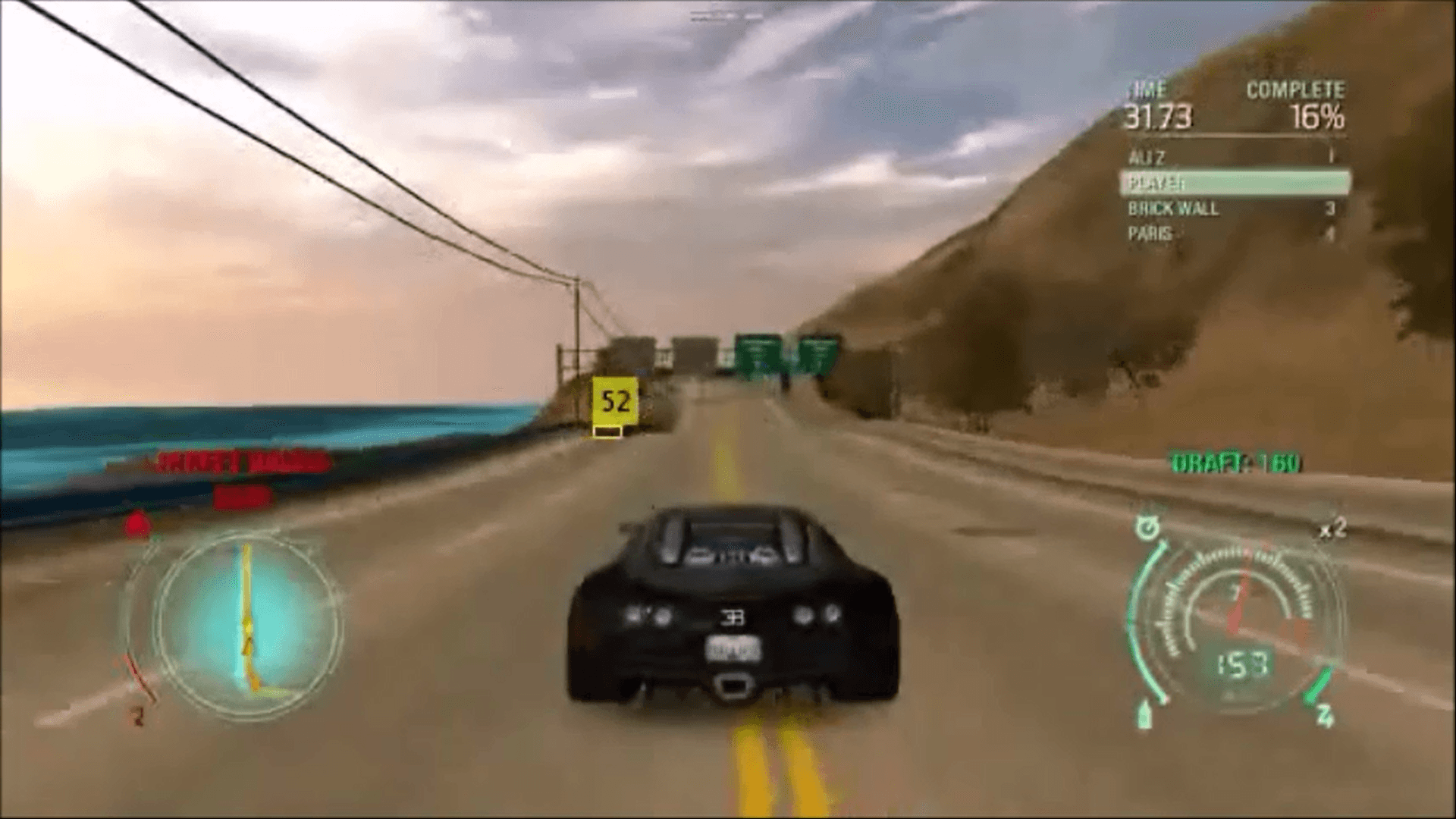}
\end{subfigure}
\caption{Kalman tracking attempts using MATLAB on a recorded video of Need For Speed game}
\end{figure}

MATLAB Embeddded Coder was not converting code properly which resulted in errors while running on Beaglebone Black board (\ref{Appendix 8}a) and function-call Coder.ceval and declaration using Coder.extrinsic for C/C++ functions were supposed to be run (\ref{Appendix 8}b)(\ref{Appendix 8}c]. Thus, we discarded MATLAB Embedded Coder.
\\\\
(We had tried Kalman tracking using MATLAB as shown in \hyperref[Figure 5]{Figure 5} and it had an accuracy of about 82.86\% when the parameters like velocity-acceleration-estimate, prior-tracking frames and others were manually coded along with color-thresholding parameters for HSV. It worked well for that video; We discarded it safely as it was parametric and Embedded Coder's performance was poor)

We were assuming that we would be able to interface GPIOs with APM directly. However, we realized that the GPIOs were not real-time and a real-time software had to replace the existing software. We ended up purchasing logic-level-converter to interface between 3.3V and 5V. We observed a delay of about 10 milliseconds (GPIO was toggled for about 1000 times and the process completed in about 20 seconds). While this is not real-time, the results could have been worse as with the case of other reports some of which have reported even 25 milliseconds or so \cite{ioPort11}. We needed real-time GPIOs for both MAVlink and Ultrasonics. Thus, we were hoping to achieve that using our code for PRUs or using register based GPIO control instead of file system based GPIO control. Customising Libsoc library was effective in solving real-time GPIO problem \cite{libsoc12}.\\

A simple file-system based control was of the following type:

\lstset{breaklines=true,basicstyle=\footnotesize}
\begin{lstlisting}[frame=single][language=C++]
const char *GPIOpin = 
"sys/class/leds/beaglebone:green:usr0/brightness";

if((GPIOhandle = fopen(GPIOpin, "r+")) != NULL){
	fwrite("1", sizeof(char), 1, GPIOhandle);
	fclose(GPIOhandle);
}
\end{lstlisting}

From this code, it is observable that the software is polling the file-system at two instances. This greatly slows down system performance. Firstly, the system is polling the GPIO file to check if the file is being accessed by some other software or if it is safe to modify the file. Then, the write process begins by opening the file and writing 1 to the corresponding register and confirming that 1 has been written (and polling the system). This was the code that we had written before attempting to integrate Libsoc library which by-passes the Linux kernel'’s timer and writes to the registers directly \cite{timeDelay13}.
\\\\
Customised code was of the form:
\lstset{breaklines=true,basicstyle=\footnotesize}
\begin{lstlisting}[frame=single][language=C++]
if(file_write(current_gpio->value_fd, gpio_level_strings[level],1)<0)
{return EXIT_FAILURE;}
\end{lstlisting}

As a result, to analyse the results, we ran the GPIO-toggle code that we had written for a million times and it took about 7 seconds (thus resulting in the switching-speed of 3.5 microseconds and boosting the speed by about 3000 times) which is coincidentally similar to Xenomai's timing \cite{beagleboneCookbook14}. As Linux kernel's timer wasn't working well (it was returning 0 for switching time), we instead designed software counter with calibration to accommodate for custom software-timer.

\subsubsection{OpenCV, image-processing and integrated machine-learning}

For image-processing application of TraQuad, OpenCV turned out to be a great-fit for real-time processing and the extensive set of software-libraries built into it. OpenCV supports programming languages of Python, C/C++ and Java (This option gave us the freedom to write code and we inevitably tried all four languages) while supporting operating system softwares of Windows, Linux, Android, Apple OSes, some of BSD linux variants and even Maemo (briefly, any standard OS supporting compilation of library) while supporting CUDA and OpenCL (These specifications were crucial for our image processing software; Lack of support of SGX graphics meant a very important consideration for embedded image processing system and so was the case with the wrapping procedure on Android which will be described) \cite{opencv15}. OpenCV also has a great community and that community includes Stanford's Stanley car, the car that won the DARPA Grand Challenge of 2005 (\ref{Appendix 11}a) and Falkor Systems (\ref{Appendix 11}b) (These codes and our own codes (\ref{Appendix 11}c) were very helpful to us). Initially, the design was considered on the basis of image-processing. We concluded that some of the operations like converting from RGB colour-space to HSV colour-space can be processor-intensive on embedded platforms and considered running such pieces of softwares on GPU. (This was important as the parameter-free considerations made us think about the 1/6th colour tolerance which can be deduced using standard-hexagonal-approximation \cite{intel16} of cylindrical HSV which meant that we were able to threshold in HSV with 1/$6^{th}$ range for Hue, 1/2 range for saturation and value parameters; a procedure which cannot be applied easily to RGB colour-space without choosing the same range-limit for all three axes.) Thus, we programmed with HSV colour-space  with 1/2 as the initial estimate for limit. 1/$6^{th}$ range made the drone susceptible to relatively observable illumination variance. Our initial attempts were to use blob as the main reference and use the centroid of the blob (using parameter-free color detection) for machine learning and navigation while using feature matching as a method of validation. We chose to use a simplified method where the user just swipes across the screen to indicate any object-of-interest that the drone must follow and we were interested in making it work for drag-and-drop interface without users needing to worry about the algorithms or inputting parameters.
\\\\
We had thought that there might be software resources just like the one that we were developing before initiating this project. We had found none (we still haven't found any excepting ours) and we started the development of this project; One of the companies Falkor Systems that had attempted visual-tracking had shut-down (\ref{Appendix 12}a). However, their open-source software gave us majority of the confidence that we were building the right software. They had used feature-matching with color-detection with Haar and LBP training. While we didn't like the idea of tuning a drone to follow only Falkor Systems's logo (\ref{Appendix 12}b) and (\ref{Appendix 12}c) and tuning the PID specifically for such configurations (\ref{Appendix 12}d) and coding object-detection-and-tracking/SURF code for a specific logo (\ref{Appendix 12}e) (\ref{Appendix 12}f), we were grateful to them as their software helped us in verifying our software.
\\\\
We chose monocular-vision system instead of stereo-vision because of two main reasons: We felt that the algorithm must be able to detect the 2D projection of a 3D object by analysing the position and size of the blob. This would avoid extensive feature-mapping (and enhance real-time processing) and avoid the heavy-bandwidth on the USB port of Beaglebone Black.\\\\

\begin{equation}
\frac{p_1}{f} = \frac{x}{f + h} \qquad \frac{p_2}{f} = \frac{x + d}{f + h}
\end{equation}

\begin{equation}
\frac{p_1}{f}(h + f) = \frac{p_2}{f}(h + f) - d
\end{equation}

\begin{equation}
d = \frac{p_2 - p_1}{f}(h + f)\implies h = \frac{f \times d}{p_2 - p_1} - f
\end{equation}

\begin{figure}[h]
\label{Figure 6}
\includegraphics[width=\linewidth]{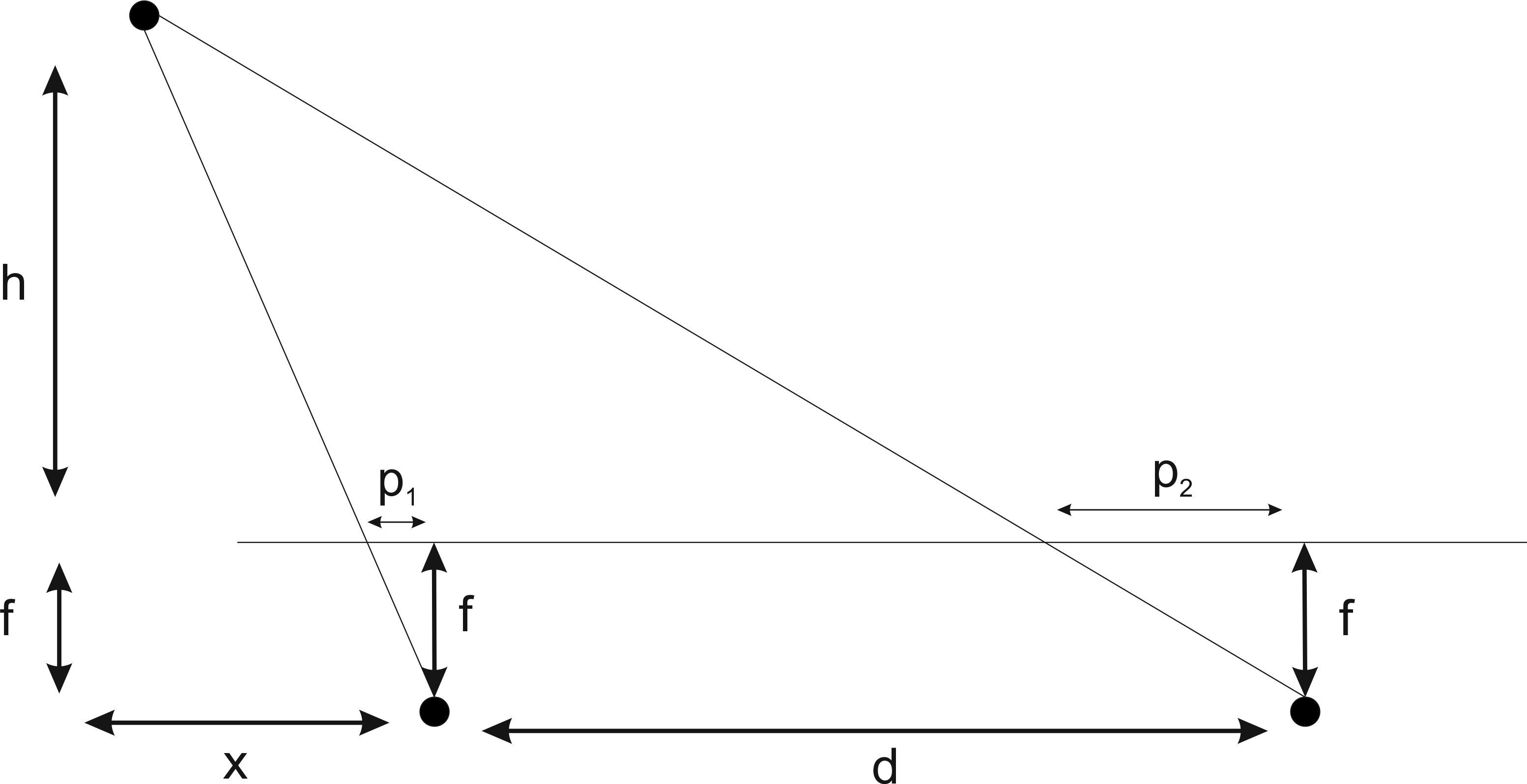}
\captionof{figure}{Stereo Vision analysis with a feature}
\end{figure}

If 'f' is assumed to be the equivalent virtual focal distance of two cameras as shown in \hyperref[Figure 6]{Figure 6}, $p_1$ and $p_2$ are the pixel-shifts of a detected feature-point and 'h' is the depth, 'h' can be calculated as shown above. This demonstrates that it is an 'embarassingly parallel' problem for GPU as a feature-point is independent of another. (Thus, we could have attempted the use of stereo-vision as an auxiliary method of validation for machine-learning algorithm. We had thought of localised depth-mapping for the generation of disparity map (processing depth of the object only by considering a rectangle of twice the dimensions of detected object around the centroid); We had thought of usage of GPU for feature-mapping and feature-detection using two levels of GPU computation (one shifted version for each camera so as to accommodate for edges on the edge of image kernels) but, we were unsure of the I/O bus of GPU and considerations of USB bandwidth made it even more difficult to capture images with two cameras.)\\

\begin{figure}
\label{Figure 7}
\centering
\captionsetup{width=\linewidth}
\includegraphics[width=\linewidth]{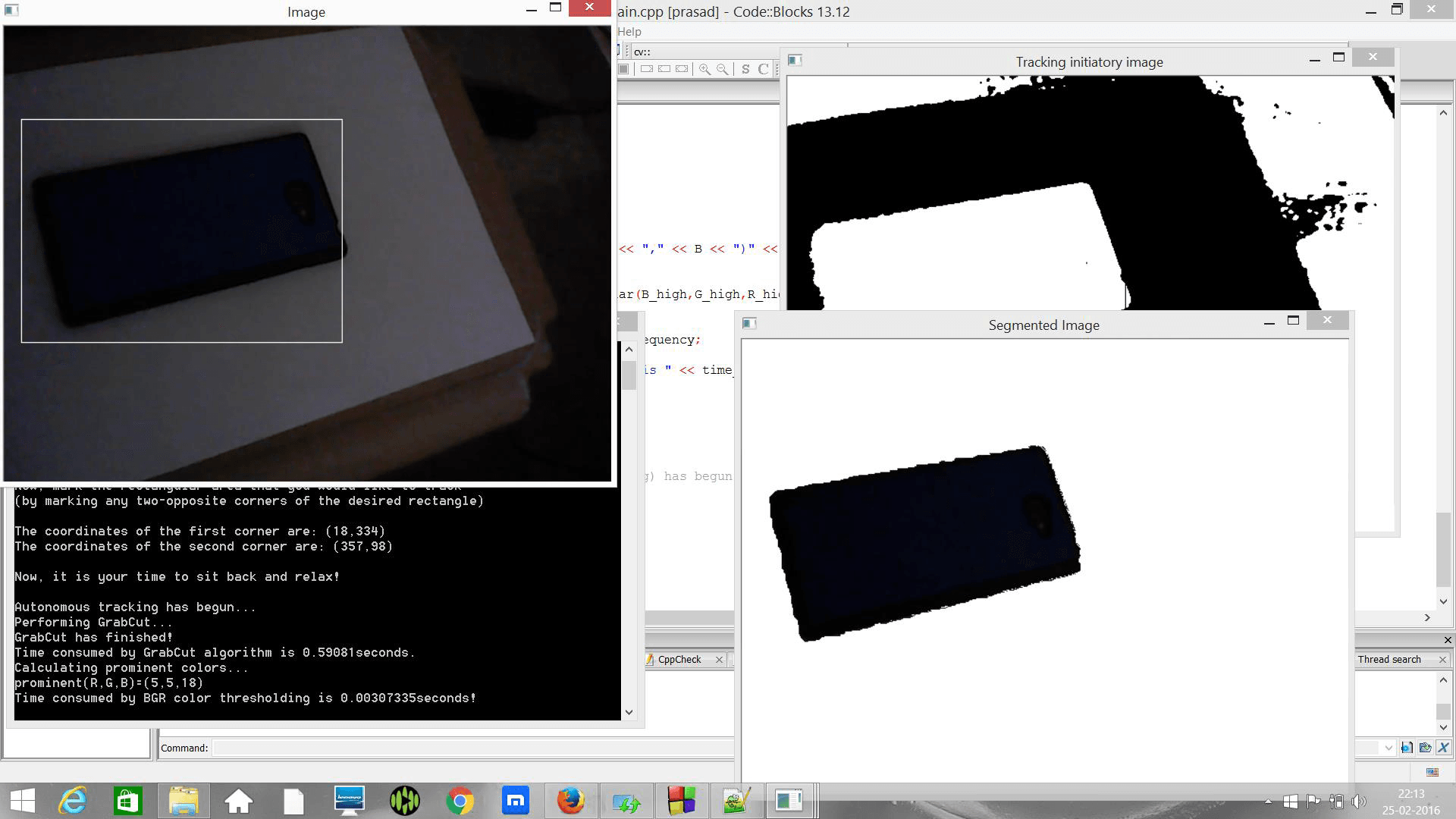}
\caption{GrabCut algorithm getting tested under low-light and other relatively severe conditions (re-created on 25th Feb 2016)}
\end{figure}

We considered GrabCut (as shown in \hyperref[Figure 7]{Figure 7}) as the Gaussian Mixture Model that is employed in it is used for both foreground and background thus making the software robust in background subtraction \cite{grabcut18}. The underlying posterior Bayesian smoothing also meant refinement of the solution around the local maximum \cite{posterioriGrabcut22}. Also, this was easier for the user; If the user didn't like the initial estimate of the object, then user was allotted to just 'swipe over' one more time. However, unlike the method of tapping-on-a-point and generating color-statistics and using connected-components or the method of contour-drawing, this method of GrabCut didn't allow the user to select as many bounding-boxes as the user likes as some of the bounding boxes overlapped. This could have meant inconvenience during the selection of a single object with relatively different shape as opposed to rectangle with horizontal orientation or selection of multiple-objects. However, when we tested, we realised that we didn't need such scenario generally. (in-fact, we had to try hard just to see if it doesn't work and it failed only in some of such cases) GrabCut performed better than what we had expected like the one shown above. Also, GrabCut was used for initiation; When that was run on BeagleBone Black, it took about 0.8 seconds to run on 320$\times$240 image. But, colour-thresholding consumed virtually zero time and the computations remained only with the calculation of centroid and bounding-box size for each image after acquiring the prominent colours (and calculating the colour-tolerance limits). GrabCut was used to compute prominent color and then, the resultant color-mask was used only inside the bounding-box (as the object of interest resided only inside bounding box). SIFT, SURF (FLANN matcher) and ORB (Brute-force Hamming matcher and Normal Hamming Matcher; Got better results with Brute-force Hamming) were tested using OpenCV 2.4 (with SURF working well on scaling invariance and was slightly quicker) and centroid was calculated using the location and the confidence-factor of features.\\

Some of the notable attempts that we undertook for image-processing so as to be theoretically sound and practically applicable can be summarised into Kalman Filter and evolutionary neural networks after which, we designed our own algorithm specifically meant for quad-rotors.\\

We had considered Kalman Filter to boost computational efficiency if needed (althought the methods mentioned above were robust enough). So, searching for literatures yielded a result on automated-tuning of Kalman filters \cite{discriminative19} and we thought that some methods were applicable to our problem as well. The example considered in discriminative training is that of a 2D GPS which resembled our problem where four corners of the detected rectangle (or radius of circle) and centroid were unreliable for Kalman filter if image-processing were to have been done with inter-frame sparse-encoding.\\

Using the authors' conventions (Discriminative training), we can use the non-linear forms:
\begin{equation} \label{eq:nonLinear}
x_t = f(x_{t-1},u_t) + \varepsilon_t; \qquad z_t = g(x_t)+\delta_t
\end{equation}
With the approximation of linearisation, we get equations of the form:
\begin{equation} \label{eq:linearApproximate}
\begin{split}
f(x_{t-1},u_t)\approx f(\mu_{t-1},u_t) + F_t(x_{t-1} - \mu_{t-1}); \\ g(x_t)\approx g(\mu_t) + G_t(x_t - \mu_t)
\end{split}
\end{equation}
If the equations consist of Jacobians or Hessian matrices with constraints like $\delta_t$$\gg$0, the linearisation approximation fails for Taylor series or other equivalent forms. This wasn't the probable case with ours. Thus, the predicted mean and covariance (and thus Kalman gain) are deduced and updated.
\begin{equation} \label{eq:meanCovariance}
\begin{split}
\bar{\mu}_t = f(\mu_{t-1},u_t); \qquad \bar{\Sigma}_t  = F_t\Sigma_{t-1}F_t^\mathsf{T}; \\ K_t = \bar{\Sigma}_tG_t^\mathsf{T}(G_t\bar{\Sigma}_tG_t^\mathsf{T} + Q)^{-1}
\end{split}
\end{equation}
\begin{equation} \label{updation}
\mu_t=\bar{\mu}_t + K_t(z_t - g(\bar{\mu}_t)); \qquad \Sigma_t = (I - K_tG_t)\bar{\Sigma}_t
\end{equation}

In the EKF learning phase, we have the following equations:
\begin{equation} \label{EKFlearning}
y_t = h(x_t) + \gamma_t
\end{equation}

The resultant joint-probability is of the form:
\begin{align} \label{jointProbability}
\begin{split}
p(x_{0:T},y_{0:T},z_{0:T}|u_{1:t}) =\\ p(x_0)&\underset{t=1}{\abovePi}p(x_t|x_{t-1},u_t)\underset{t=0}{\abovePi}p(y_t|x_t)p(z_t|x_t)
\end{split}
\end{align}
where,
\begin{equation}
\begin{split}
p(x_t|x_{t-1},u_t) = \mathcal{N}(x_t;f(x_{t-1},u_t),\textit{R}); \\ p(y_t|x_t) = \mathcal{N}(y_t;h(x_t),\textit{P}); \\ p(z_t|x_t) = \mathcal{N}(z_t;g(x_t),\textit{Q})
\end{split}
\end{equation}
However, \textit{h} is assumed to be a linear projection operator which wasn't the case with the described problem. The resultant joint-probability wasn't linear-Gaussian as expected.
(This posed problems when vehicle veered along the path; But, the results achieved with this were close to 80\% in specific cases and enticed great hopes and we are grateful to these authors for the insight)\\

NEAT algorithm solves multi-dimensional machine learning problems by generating neural network layers while effectively avoiding redundant networks while employing inheritance and crossovers \cite{neat21}. But, we found that this method had parameters for mutation, sigmoidal thresholding function and more and so is the case with HyperNEAT. While these parameters can be tuned using successive parameter-free machine-learning algorithms, we were unsure about the development of it given our limited database. (This is a potential future work that we are considering)\\

\begin{figure}[h]
\label{Figure 8}
\centering
\begin{minipage}{.5\linewidth}
  \centering
  \includegraphics[width=0.95\linewidth]{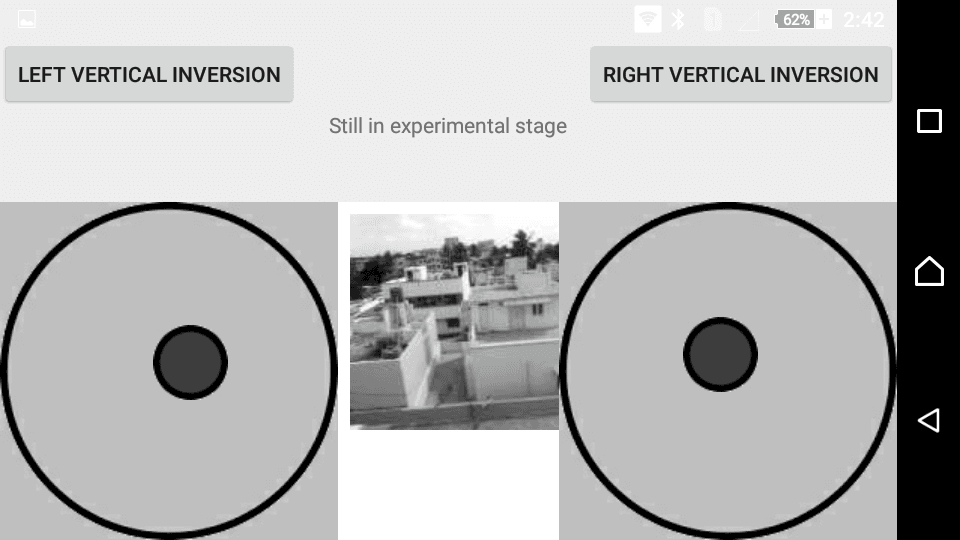}
  \label{pro_joypad}
\end{minipage}%
\begin{minipage}{.5\linewidth}
  \centering
  \includegraphics[width=0.95\linewidth]{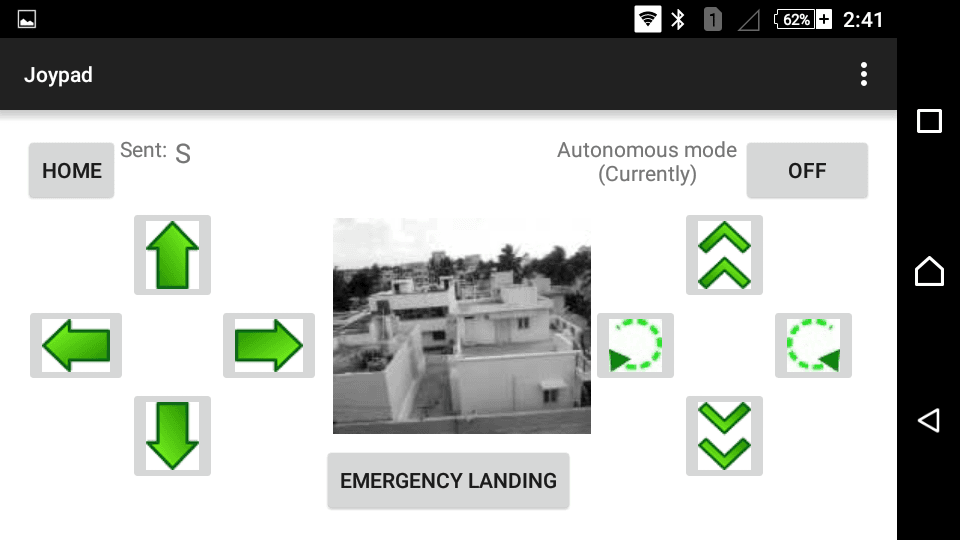}
  \label{simple_joypad}
\end{minipage}
\captionsetup{justification=centering}
\caption{Pro-joypad mode on left and Simple-joypad mode on right}
\end{figure}

The image-processing software itself was developed with top to bottom approach by considering the hardwares used as well. We thought that the analysis of the orientation for the object-tracking using homography and perspective transforms were unnecessary for the first-prototype (before flying with Beaglebone and testing image-processing algorithms). (RANSAC is known to be computationally intensive as well) We needed a tracking system for a rigid-body with an additional capability of tracking multiple states of rigid-body (as in human hand moving around while walking). We were trying a lot of algorithms and checking by manually moving the camera before attaching to the drone; Most of the algorithms were relatively computationally intensive on embedded systems (including TLD which works relatively well in long-term detection among the standard algorithms) or were not robust enough. Searching for the research literature reinforced that belief \cite{2011_23}\cite{2006_24}\cite{2014_25}. We thus believed that the use of frameworks like OpenCog (\ref{Appendix 16}) for associative learning (image processing along with audio) may complicate the problem or defy the problem statement itself and so was the case with ROS \cite{ros26} which we then considered for it's amazing simulations \cite{rosSurvey27} (After reading research literature, our thumb rule turned out to be OROCOS for real-time industrial robots and ROS for general purpose robotics). Thus, we ended up using a combination of colours and features and determining the position and size of the projection of the object. We also included the motion of the camera itself to accommodate the variation in image properties. For the detection of features, a minimum number of 3 non-collinear features had to be selected in 2D plane; We used this as a condition to determine occlusion or the quality of detection. We rejected the idea of localised template matching with least mean square error condition due to scaling and rotation variance. We used the default parameters of APM as it is supposed to be meant for general purpose (\ref{Appendix 15}). We didn't use Autotune feature of APM as it can cause ESC-transient issues and even burn ESCs (\ref{Appendix 21}f) as the AutoTune can request very large and very fast changes. While we were able to get the 2D tracking, for third dimension, we used the size of the blob (Root Mean Square radius) as the reference for third dimension. Reading the standard documentation, it can be understood that the flight-controllers generally refresh at the rate of 50 Hz (\ref{Appendix 16}a) and the PWM time ranges from 1-2 milliseconds (\ref{Appendix 16}b). We ended up considering an option of one-second-calibration which incrementally iterates from zero to maximum value of a parameter and detect bias and record flight-data to automatically tune parameters if needed (as in increasing the throttle incrementally within one second and tuning PID according to scenario; the process can then be repeated for yaw and pitch instead of roll (\ref{Appendix 16}c); But, we observed significant yaw-wobbles during testing and thus refrained from using this method although it worked great for pitch and throttle). For the controller part of the phone, three modes were designed: Simple, Pro-joypad and autonomous modes. The simple mode (as shown in \hyperref[Figure 8]{Figure 8}) was meant for those users who would fly drones without much training; the drone would then use automatically tuned parameters for it's stabilisation. Multi-touch pro-joypad mode (as shown in \hyperref[Figure 8]{Figure 8}) was meant for professional users and autonomous mode was used to display the live-streaming information.

Initially, we had thought of usage of two cameras that would 'switch' on the basis of movement of the object. The orientation that was thought of was a horizontal one for one camera and another one focusing vertically downwards. We realised that the USB bandwidth would pose a significant problem during 'switching duration' when both cameras work; We had also known about the rotational transformations that had to be applied if projection angle was not 0$\degree$. We ended up retaining only the horizontal camera (and thus avoided minor pitch compensation using STATUS; pitch compensation was not possible for vertically downward facing camera if the object moved relatively quickly such that the displacement in the image was more than 1/$4_{th}$). 0$\degree$ also meant an inherently stable drone; But, in such a setting, the object's height must be greater than drone's dimensional height (not necessarily altitude) or if the object is smaller, then, the drone must fly just above ground (using LIDAR or Laser-ranger if needed). (1-2 metre from the ground has been mentioned in Altitude Lock Documentation of APM (\ref{Appendix 24}))\\

We then used online on-board only training method STATUS which we had developed. Although off-board computation methods like internet can be used along with GPS while employing STATUS, we practically found the unnecessary latency-compensation-software that is needed for internet (GPS can provide better flight-stability and avoid popularly known 'toilet bowl' effects at lower speeds; But, we didn't test with GPS). It is a method where the angular velocity of yaw and throttle are controlled by the linear equations while considering non-linear effects in $3^{rd}$ dimension. We chose velocity and not acceleration or any further derivatives as it may end up introducing transients or end up losing track of the object if image-processing is not real-time. We use simplified representations to describe equations. \textit{yaw} represents magnitude of angular velocity of yaw, \textit{throttle} represents the magnitude of velocity of throttle, \textit{width} and \textit{height} represent the width and height of the image, \textit{x} and \textit{y} are the horizontal and vertical distances from the left-top corner of the image. \textit{forward} represents horizontal, forward velocity.

\begin{equation}
\begin{split}
yaw = 1 - [(step(x - \frac{width}{4}) - step(x - \frac{3\times width}{4}))\\\times(1 - \frac{4}{width} \times modulus(x - \frac{width}{2}))]
\end{split}
\end{equation}

\begin{equation}
\begin{split}
throttle = 1 - [(step(y - \frac{height}{4}) - step(y - \frac{3\times height}{4}))\\\times(1 - \frac{4}{height} \times modulus(y - \frac{height}{2}))]
\end{split}
\end{equation}

\vspace{-8mm}
\begin{minipage}{0.5\linewidth}
\begin{gather*}
\forall modulus(t) = 
\begin{cases}
-t, \qquad t<0\\
t, \qquad t\ge 0\\
\end{cases}
\end{gather*}
\end{minipage}
\begin{minipage}{0.5\linewidth}
\begin{gather*}
\& step(t) = 
\begin{cases}
0, \qquad t<0\\
t, \qquad t\ge 0\\
\end{cases}
\end{gather*}
\end{minipage}

\begin{figure}[h]
\label{Figure 9}
\centering
\includegraphics[width=0.5\linewidth]{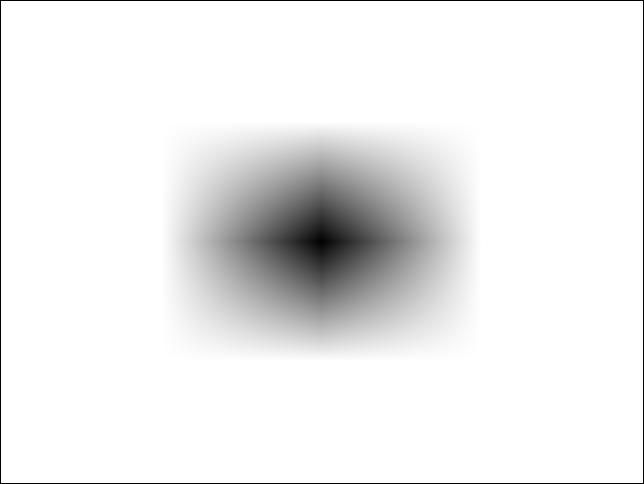}
\footnotesize
\flushleft
The diagram above depicts the normalised power distribution. White represents maximum value of the aspect (\textit{angular velocity of yaw} and \textit{throttle velocity}; \textit{yaw actuation velocity} = \textit{maximum yaw} $\times$ \textit{normalised yaw angular velocity} and \textit{throttle actuation velocity} = \textit{maximum throttle} $\times$ \textit{normalised throttle velocity}) and black represents 0. The depicted normalised magnitude has been computed by the equation normalised $\overline{\overline{\textit{angular velocity of yaw}} \times \overline{\textit{throttle velocity}}}$ (complementary probabilities).
\caption{Normalised magnitude of composite proportion of throttle along vertical axis and yaw along horizontal axis for a given amount of power. (Except the border)}
\end{figure}

Yaw's equation and throttle's equation accounted for variability in bounding-box size as well and linearisation across the entire axes has been avoided (as shown in \hyperref[Figure 9]{Figure 9}) after testing several possibilities for effective yaw-control on-field. \\

For pitch control, we started with the basic assumption of statistics: Lesser $\sigma$ means higher speed for similar deviations along other dimensions.\\
\\$df(x) = -\frac{Area_{dx}}{variance}$ where \textit{x} is the one-dimensional statistical input.\\\\
Thus, $df(x) = -\frac{f(x)xdx}{\sigma^2}$ has solution $f(x) = ce^{-\frac{x^2}{2\sigma^2}}$. ($Area_{dx}$ decreases as \textit{x} is farther from mean which results in negative slope) Thus, normalised Gaussians were used to control the velocity along the third dimension\ref{Appendix C}. (We did not start with model; But, we ended up with Gaussian curve as a result)

\begin{equation}
\sigma_{centroid} = \sqrt{\frac{\Sigma_{0}^{n} (distance_{image})^2}{n}} 
\end{equation}
$\forall$ $distance_{image}$ = distance of the centroid of the blob from the center of image $center_{image}$ ($\frac{width}{2}$,$\frac{height}{2}$)
\\
\begin{equation}
distance_{image} = \sqrt{(x - \frac{width}{2})^2 + (y - \frac{height}{2})^2}
\end{equation}
\\\\
($0^{th}$ frame indicates initialisation; If this needs to be avoided, a Gaussian model with $\sigma$ = $\frac{width}{4}$ with Gaussian-mean = $center_{image}$ produces equally nice results practically)\\

$\sigma_{centroid}$ was calculated for distances of centroids from $center_{image}$. It was calculated to accommodate for the varying conditions; If the pixel-distance of the new centroid from the centre of the image was more than the standard-deviation, we would retain the existing Gaussian model and consider this outlier-centroid for the calculation of standard-deviation so as to consider further updates of Gaussian model. This can contribute significantly to long-term machine learning as we did not have much problems with memory. However, if the learning needs to be restricted to a certain number of frames, the algorithm can then be made to compute for standard-deviation for every frame since the previous second or the number of frames retained can be equal to the current standard-deviation. (We faced this problem on Android which will be discussed) For non-binary-matchers like Euclidean-distance for feature-matching metric were iteratively increased by 2 until 3 features were got and decreased by 1 if there were more than 3 features for every image captured. For binary-matchers, (ex: Hamming distance) binary-metric distance was increased and decreased to maintain minimum error.

\begin{equation}
\footnotesize
\textit{forward velocity} = \frac{maximum \thinspace forward \thinspace velocity}{\sigma_{centroid}\sqrt{2\pi}}\exp(-\frac{1}{2}(\frac{distance_{image}}{\sigma_{centroid}})^2)
\end{equation}

An interesting extension of this was the normalised division of path-planning weights of the algorithm that we had designed for 2D (built for GNU Octave (\ref{Appendix 19}) and tested in MATLAB\footnote{This path-planning software was created and ideated by Prasad N R even before the beginning of development of TraQuad}) \cite{pathPlanning29}. This was relatively easy in 3D as the normalised probability had to be applied across three dimensions (x,y,z) instead of two (x,y) thus having a euclidean-distance function mapped to normal probability while the augmented matrix's number of dimensions remained the same. The sum of the entire path was stored and for all the summations of the respective entire paths, the highestSum/currentSum and multiplying this ratio with the Gaussian velocity as long as the path-planned-velocity is lesser than the maximum linear velocity parameter of APM.

\subsubsection{Software simulations}

Our aim was to simulate a quadcopter with a horizontally aligned camera and Lidar (only for validation) and a rover simultaneously which represents moving object using a general purpose simulator. We found that ROS and Gazebo is the best software combination and many other SITL simulators (\ref{Appendix 21}e) including CRRC simulator which is not a standard simulator \cite{rosSurvey27}\cite{rosGazeboSurvey28} were then carefully assessed and while there are flaws (\ref{Appendix 28}), we conclusively found that ROS and Gazebo is a great combination; But, there were no such object-tracking simulation in ROS-Gazebo and we chose to build one. (Note: As of 21 January 2018, we have come across another ROS Gazebo simulator. But, we believe that the installation instructions are not detailed and the scene used does not represent the complete object tracking drone's scenario which should involve tracking of even the toppling of rover when used on a challenging track. Also, the ARDone simulation is great for general drone simulation. But, for our case, it might not represent the Beaglebone black simulation directly\cite{rosGazeboVisionSimulator}) (FlightGear simulations did not work out very well for rover physics \hyperref[Appendix 14]{Appendix 14})\\

\begin{figure}[h]
\label{Figure 10}
\centering
\includegraphics[width=\linewidth]{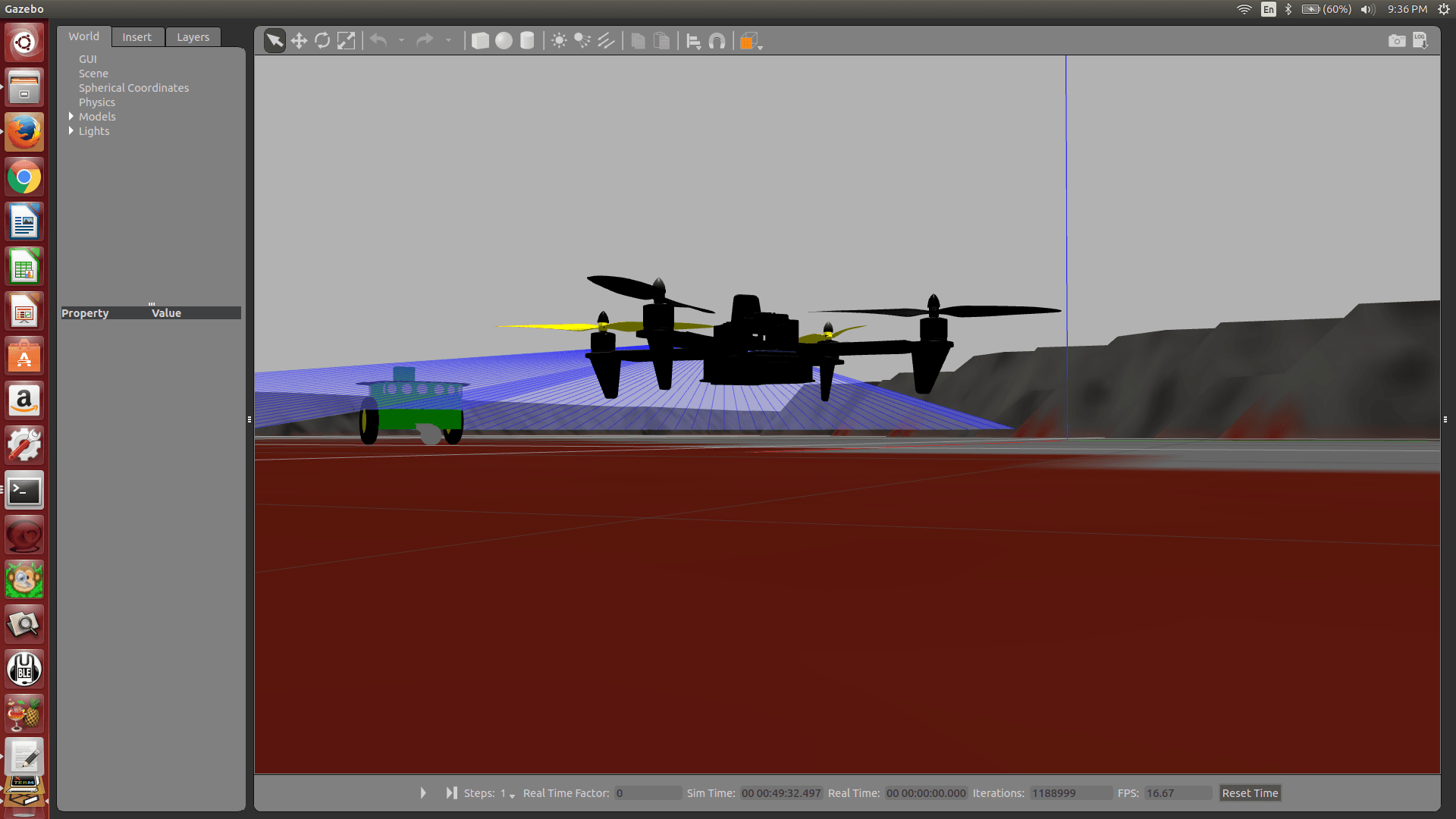}
\caption{Actual tracking is being tested out.}
\end{figure}

The only common operating system for both Gazebo (which supports Mac and Linux) and ROS (Linux only) is Ubuntu Linux (\ref{Appendix 17}). On a single Lenovo y50-70 laptop with Ubuntu 14.04, ROS Indigo and Gazebo7 have been installed and software-simulation has been performed for both car and quad-copter. We haven't used ROS2 as it is still in development (\ref{Appendix 21}c) and the versions have been checked by attempting multiple versions, attempting to solve multiple compilation errors (\ref{Appendix 21}a). Erle-Gitbook mentions the compilation errors which hasn't compiled yet and we needed a stable simulator (\ref{Appendix 21}b) and we used Erle-copter and Pioneer 3DX models; We have chosen Erle-copter as it indirectly uses BeaglePilot project (and thus validates Beaglebone's algorithms; As of 9th October 2016's repositories). Thus, we effectively tried to use APM flight testing simulation in Windows MAVproxy SITL (not stable) and chose custom ROS Gazebo SITL for algorithm-testing (\ref{Appendix 21}d). We didn't use HITL or test the APM with SITL as we were using default parameters along with arming checks. Out of Throttle, Pitch, Yaw and Roll commands, we eliminated Roll command as we needed two angles and one linear magnitude to represent three-dimensional aero-dynamic force vector.\\

\begin{figure}[h]
\label{Figure 11}
\centering
\includegraphics[width=\linewidth]{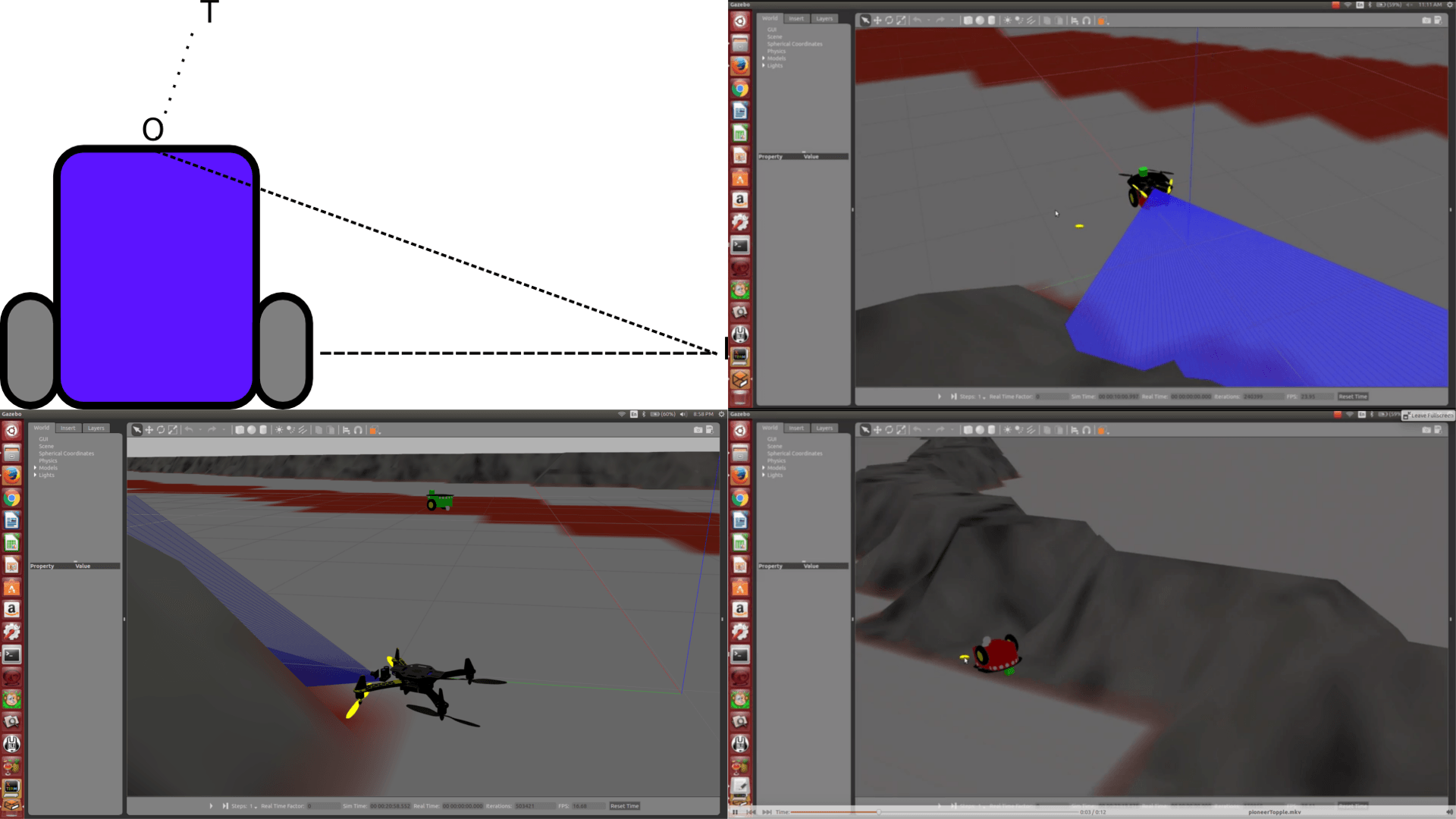}
\caption{Left top: Ackerman or wagon steering, Right top: physics testing by landing copter on rover, Bottom: Intensive physics and algorithm testing resulted in Copter and Rover toppling over.}
\end{figure}

Our setup includes a Pioneer 3DX (whose chassis color has been changed) following the line with 64$\times$48 resolution camera at 5Hz(\ref{Appendix 27}). This model is about the size of copter itself and is challenging as the height of the object to be tracked must be equal or greater than the size of the copter\footnote{http://ardupilot.org/dev/docs/beaglepilot.html}. 
Erle-copter has 10Hz Lidar and 320$\times$240 resolution camera running at 5 Hz. Pioneer 3DX has a normalised angular velocity of 1rad/s while the maximum stable velocity of this relatively small rover is about 0.5 m/s (after this, it loses track and eventually topples and we have even run at 0.75 m/s) and thus, we set it at 0.5 m/s for benchmarking. The correlation between physics of our simulator and reality seemed to be strong. This has been shown in \hyperref[Figure 11]{Figure 11}. The resultant setup is as shown in \hyperref[Figure 10]{Figure 10}.

We were trying to simulate pure-Ackerman steering or fifth wheel (axle articulated) steering with each wheel's tyre contacting ground at only one point with two-wheel drive with each wheel's angular velocity being determined by ratio of radius from intersection point I and radius of wheel (Not rear wheel steering or crab-steering or frame-articulated steering)\cite{steering}. The orthogonal imaginary line intersected with axle's axis gives the radius of steering when camera is mounted at O and T is the target steering location. But, it requires that steering angle must be less than 90 degrees which paved way for complicated algorithms (we were trying on Erle Rover and FDM physics itself consumes computational power) and the mechanical reverse of Rover in software was not eaily programmable.\\

We faced problems in Yaw in OverrideRC method and while this was not a major issue in hardware, SITL showed significant oscillations\cite{yawProblem} (which we believe are because of ODE Physics engine updation which requires more powerful computers; We also observed non-linear software effects when performance of computer was greater than about 50\% of CPU capacity). Thus, we had to slightly modify STATUS to avoid inter-frame image processing and include hard-coded Gaussian at the centre of image and consider the variance as the area of blob and the Override RC values were from 1400 to 1600. (a problem which we believe can be easily solved if we use more powerful computer; An additional extension to STATUS can be the consideration of $\sigma_{color}$.\\

\begin{equation}
\frac{\sigma_{color}}{3\times256} = \frac{\sigma_{centroid}}{\sqrt{(\frac{width}{2})^2 + (\frac{height}{2})^2}}\Bigg|\sigma_{color} \ge 1/6^{th} tolerance-limit.
\end{equation}

\subsubsection{USB}

\begin{figure}[h]
\label{Figure 12}
\centering
\includegraphics[width=\linewidth]{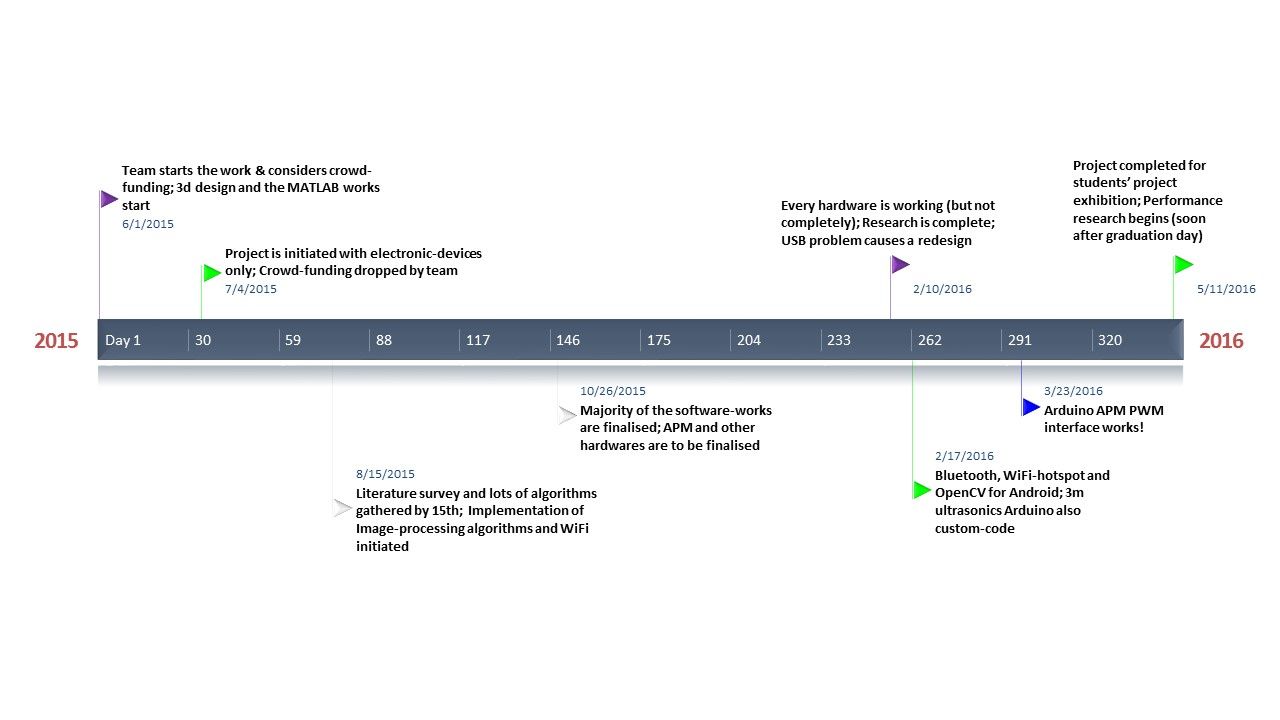}
\caption{Timeline of development of TraQuad}
\end{figure}

Beaglebone Black has one USB (host) port and we were using that port along with a hub for camera and WiFi (along with data-less upstream power-bank). But, our tests with the camera demonstrated that the USB was running in 12 Mb/s Full speed instead of 480 Mb/s High speed as mentioned in the data-sheet. We thought that the change in the speed coupled with the introduction of isochronous mode instead of bulk-transfer mode would ease the data-transmission via USB. While searching for solutions, we happened to glance through some of the USB driver installation issues (\ref{Appendix 18}a) which had warnings of the EEPROM getting locked and corrupted. As Beaglebone's datasheet mentions FTDI devices, we read about the recovery of the USB-EEPROM and confirmed that the software can link USB and EEPROM and damage EEPROM (\ref{Appendix 18}b). The disaster recovery document mentions about users occasionally corrupting the EEPROMs and rendering the module unusable and about the possibility of combinations which can render the device non-enumerable. CAT24C256 EEPROM chip mentioned in Beaglebone's datasheet has about 1 million program/erase cycles and 100 years data-retention-time as per the datasheet (\ref{Appendix 18}c) and D2XX programmer was the only option left for that and reading the documentation consumed relatively much time (\ref{Appendix 18}d). Otherwise, modifying/directly ordering some of the devices containing those chips was the only option left (but, we couldn't find USB 2.0 high-speed device properly) (\ref{Appendix 18}f). The programming guides are filled with warnings (\ref{Appendix 18}e) which made us consider the option of considering Beaglebone or working on a new device altogether.
The timeline is as shown in \hyperref[Figure 12]{Figure 12}.

\section{Redesign}

\begin{figure}[h]
\label{Figure 13}
\centering
\includegraphics[width=\linewidth]{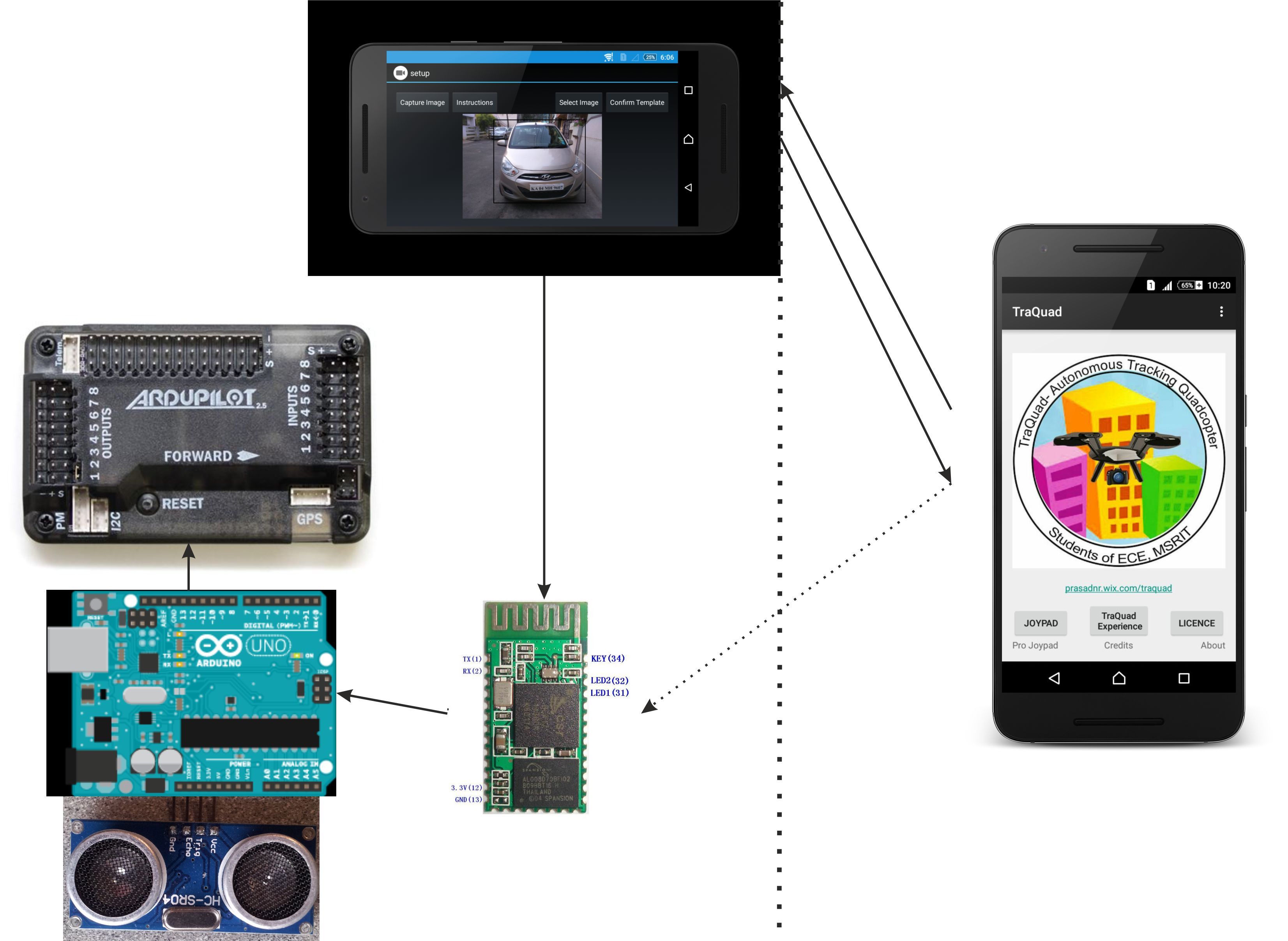}
\caption{Final Architecture}
\end{figure}

\begin{figure}[h]
\label{Figure 14}
\centering
\begin{minipage}{.5\linewidth}
  \centering
  \includegraphics[width=0.95\linewidth]{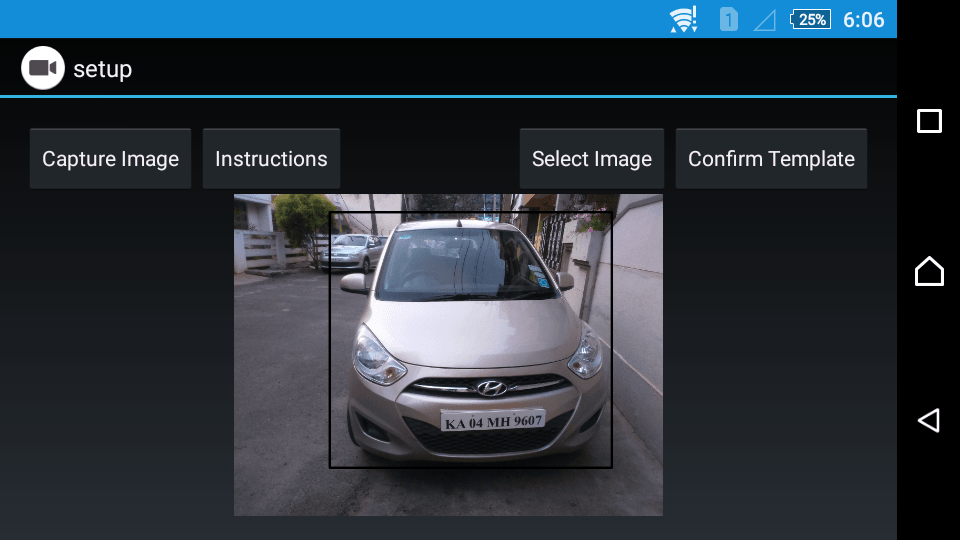}
  \label{drag_and_drop}
\end{minipage}%
\begin{minipage}{.5\linewidth}
  \centering
  \includegraphics[width=0.95\linewidth]{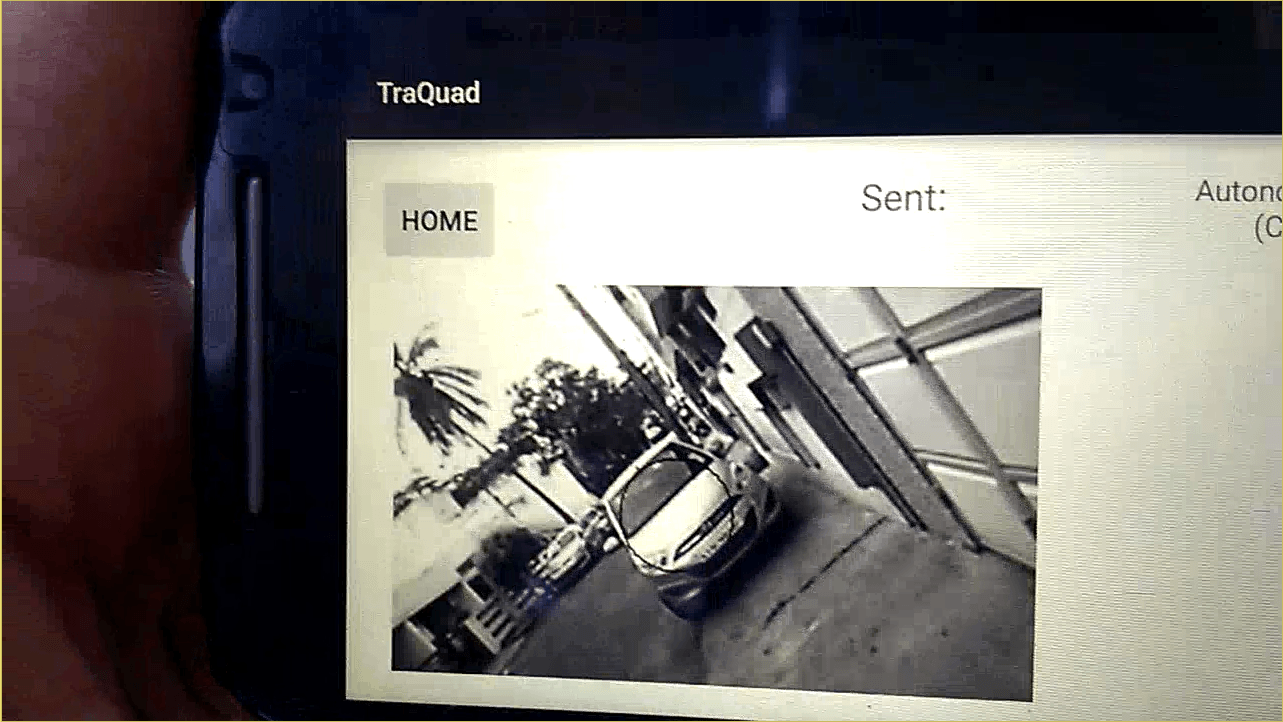}
  \label{tracking_mode}
\end{minipage}
\caption{Left: Drag-and-drop interface for tracking initiation; Right: Tracking-mode}
\end{figure}

We then resorted to the usage of Android phone for image-processing also. The resultant architecture is as shown in \hyperref[Figure 13]{Figure 13}. The interfaces are shown in \hyperref[Figure 14]{Figure 14}. The Android phone captured NV21 images which was then converted to RGB. WiFi and Bluetooth were used as the interference between WiFi's Direct Sequence Spread Spectrum and Frequency Hopping Spread Spectrum is negligible (\ref{Appendix 22}). OpenCV library was linked with Android application and proprietary OpenCV manager containing C/C++ codes were used as NDK was not easy to optimise and work with. MJPEG streaming was achieved by using and modifying Seeed Studio's Webcam Android application \cite{SeeedStudioWebcam30}. While modifying, we had to cautious about two aspects: Big-little architecture of Android \cite{bigLittle31} and Garbage-collection deployed for every software-thread as a part of automatic-memory-management by Android \cite{performanceAndroid32}. Big-little architecture meant that the phones are meant to save battery-power and not optimised for performance. Garbage-collection meant that some of the object-references and threads can be terminated if the memory consumed turned out to be relatively intensive. Arduino was used for obstacle avoidance and to control the APM. The PWM values of throttle, pitch, yaw and roll (around 1500; in fact slightly lesser than 1500 for throttle) varied between 1000 microseconds and 2000 microseconds which were generated which were mapped to 45\degree to 135\degree angles of servo as Arduino's servo library was being used (This was then adapted to work with 1400 to 1600 in our practical tests although high-speed tracking caused issues with yaw when RC values of 1200 to 1800 were used). Almost every single image-processing algorithm had to be modified to accommodate for the factors mentioned above (GrabCut, SIFT and SURF no longer worked properly without NDK). ORB was used for feature matching along with Brute-force Hamming matcher as it is an amazing alternative to SIFT and SURF \cite{ORB33}. The BRIEF's rotational invariance in ORB is nice (it was detecting even orthogonal images); But, the FAST's scale invariance is not satisfactory for real-time image-processing in Android when the variance exceeds a certain standard-deviation for a given resolution after which non-linear software effects are observed. (A notable mention: During this period, we glanced through the notable Stanford's driving software \cite{stanley34} and opted to modify our Beaglebone's working algorithms and add necessary filters or parameters wherever necessary. We observed some of the best algorithms which were directly applicable to our Android algorithms; They have discarded those objects which occupy less than 1\% of the entire image, used watershed algorithm using edges, effectively managed traffic-signal detection and more.) We continued with feature-matching; But storing images reduced the frame-rate to 1fps (Sony Xperia Dual M2); Thus, we devised Global-class which held variables in cache. The features of the template were stored in Global class which was computationally efficient. This part of the software was more reliable and we achieved about 320$\times$240@150 fps just to stream. We have used dynamic Hamming matcher threshold where the threshold-distance is increased by 2 if a feature is detected and decreased by 1 if a feature is already getting detected. Centroid was calculated by considering the locations of filtered-features and confidence of filtered-features as weights. RMS distance was calculated by considering the square-root of Euclidean-distance of all filtered-features. Features were filtered on the basis of colour; If a feature was exceeding 1/12th colour-tolerance limit on either of sides from mean colour of features, then it was discarded. Storage of many images in cache triggered garbage-collection and battery-optimised mobile was unable to perform well on set of images. Thus, we reduced the image to 9$\times$9 image-kernel by considering the location of filtered-features in image (threshold being 1/4th the total number of frames $\implies$ 10/4 or greater than 2 frames) and scaling it down. We used 10 such images and averaged the radius and centroid location. Increasing or decreasing the kernel size and number of images suited for different needs. Thus, we have achieved about 640$\times$480@25fps with 95\% detection rate and 78\% tracking efficiency with the re-designed architecture.

\section{Results}

We had used MATLAB for Kalman Filter based tracking with 40 training frames, 0.7 as minimum back-ground thresholding ratio, minimum blob-area of 50 pixels, 15 $\times$ 15 square-shaped noise-filler and 3 frames as threshold for lost-tracks while 10 frames were used to confirm for the initiation of tracking. This resulted in about 82\% tracking efficiency for NFS game while it rarely worked for practical tracking videos of our case (and fell below 45\% sometimes). This was a high-dimensional machine learning problem and we had to discard this and design controller also with our own image processing algorithm.\\

We then modified LibSoc library for register level access through C++ code and clocking at 1Ghz, we have achieved about 3.5 $\mu$s switching time (Xenomai-RTOS-software's average switching time = 3.916$\mu$s, worst-case time = 25.5$\mu$s and minimum-time = 2.75$\mu$s) and we encountered USB issues as mentioned above. We used our own controller algorithm with partial-linearisation with base-statistical belief which resulted in Gaussian control for pitch. GrabCut consumed about 0.8 seconds while color-thresholding consumed about 5 milli-seconds on BeagleBone Black for 320$\times$240 image.\\

\begin{figure}[h]
\label{Figure 17}
\centering
\includegraphics[width=\linewidth]{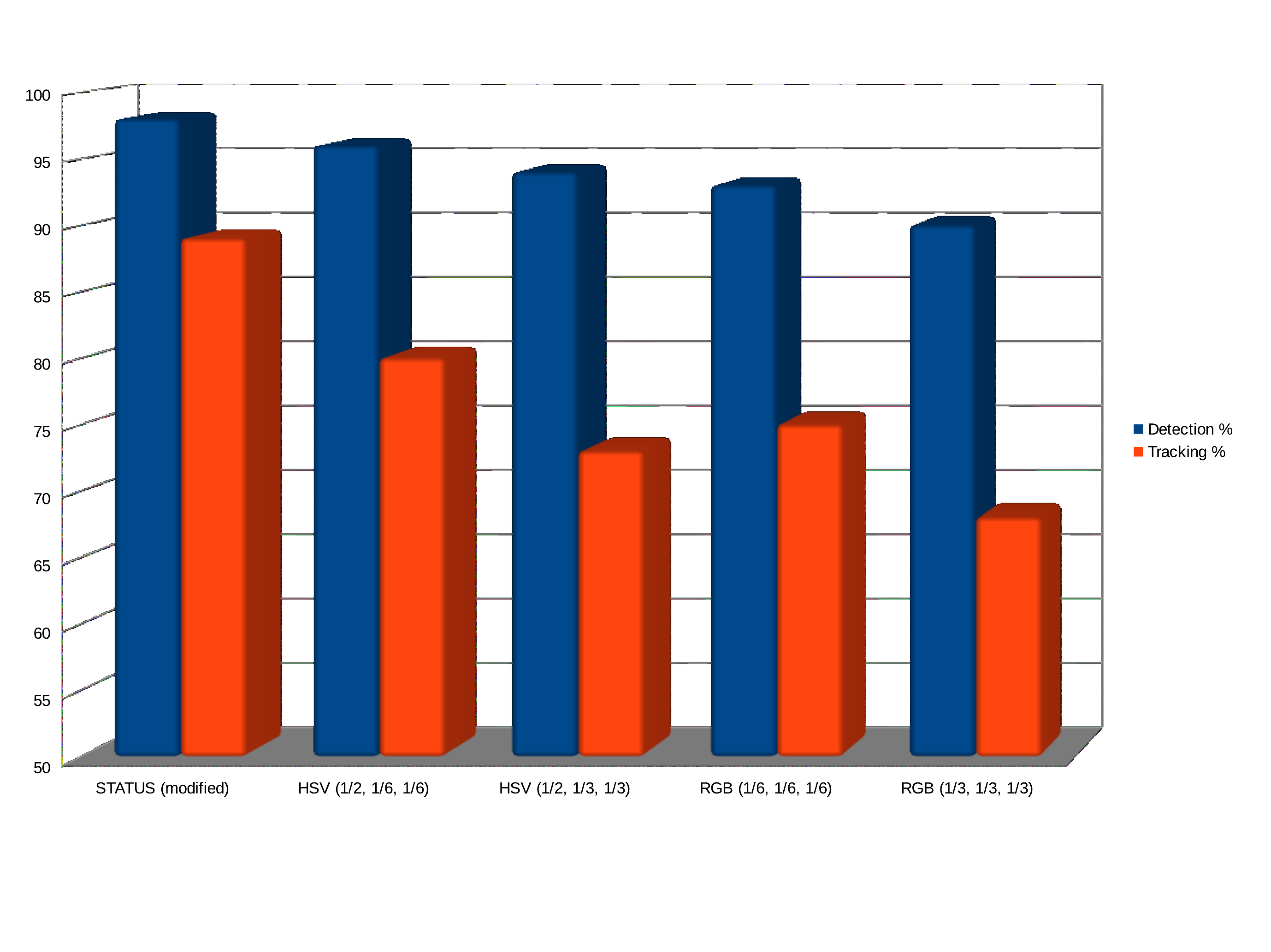}
\vspace{-15mm}
\caption{Detection and Tracking rates of various algorithms in SITL.}
\end{figure}

In simulation, the rover moves at 0.5 m/s and quad-copter is fixed with 320$\times$240 image-resolution with 5 Hz frame-rate camera. The results are as shown in \hyperref[Figure 15]{Figure 15}, \hyperref[Figure 16]{Figure 16} and \hyperref[Figure 17]{Figure 17}. (We have included the image of our test quadcopter in \hyperref[Figure 18]{Figure 18})

\begin{center}
\label{Figure 15}
\begin{minipage}{0.5\linewidth}
  \centering
  \includegraphics[width=0.95\linewidth]{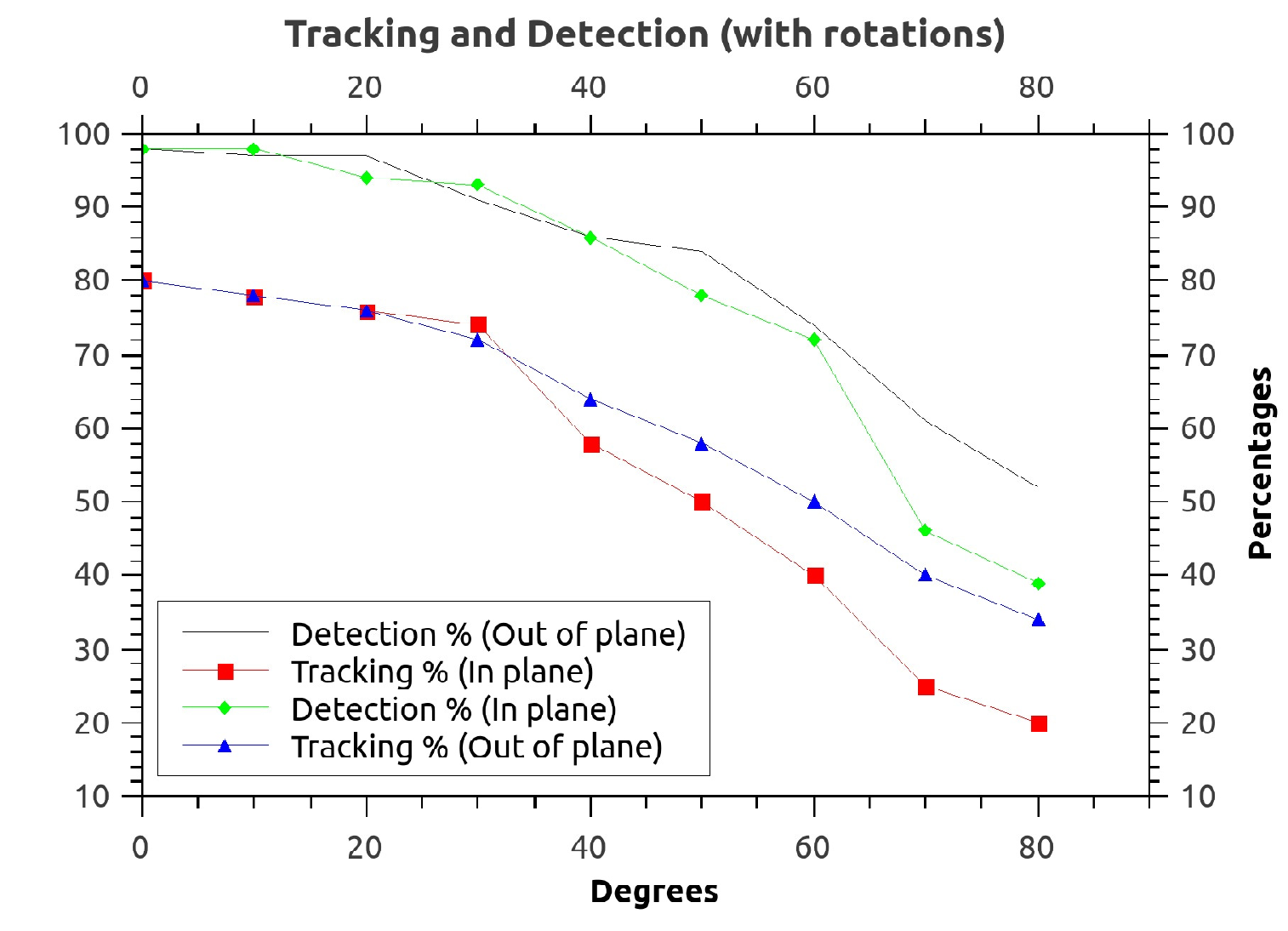}
  \label{rotation}
\end{minipage}%
\begin{minipage}{0.5\linewidth}
  \centering
  \includegraphics[width=0.95\linewidth]{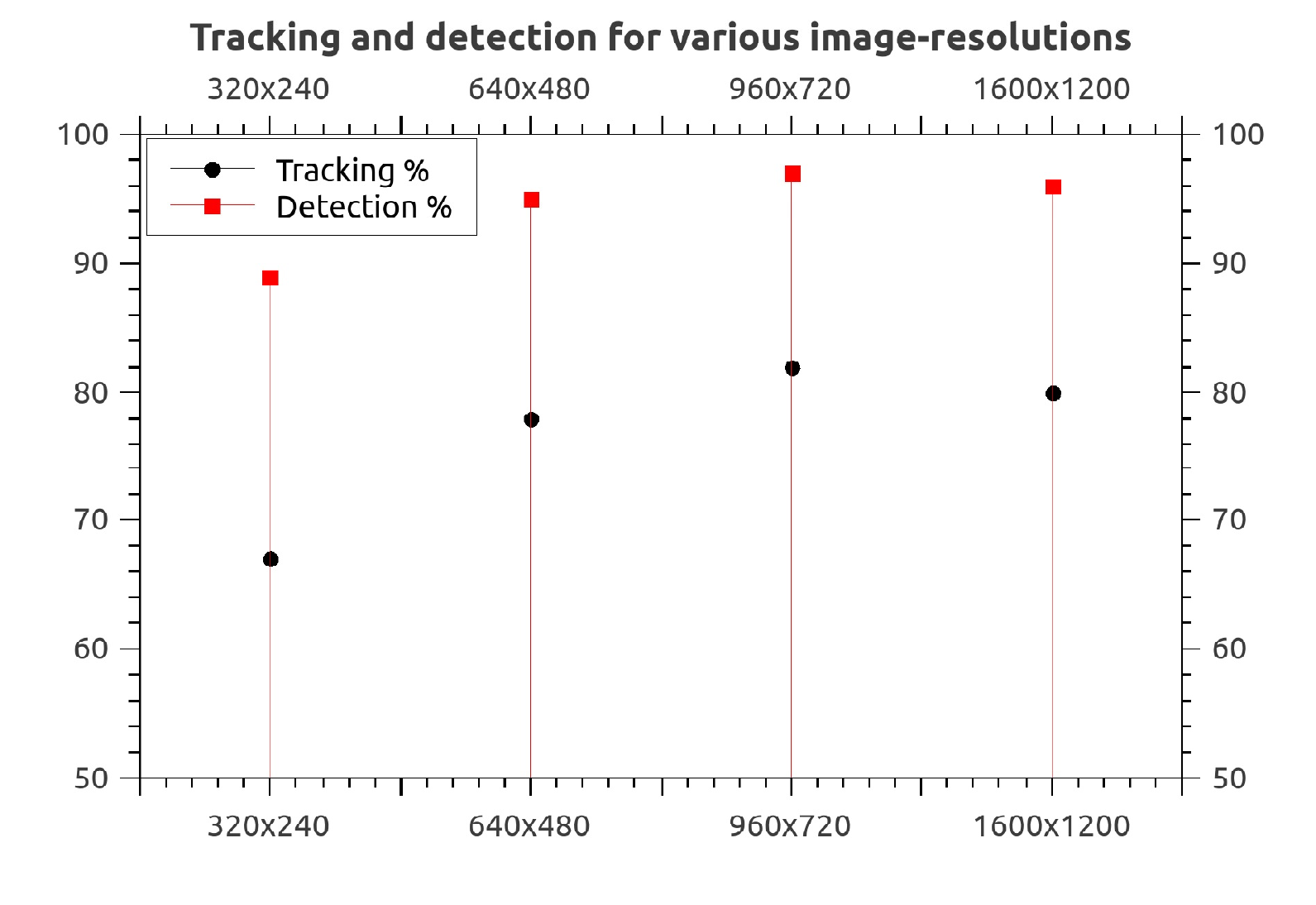}
  \label{imageResolution}
\end{minipage}
\captionof{figure}{Detection and tracking rates of various parameters with modified STATUS with actual hardware (Android). Top-left: Rotational tracking, Top-right: Tracking for different resolutions}
\label{Figure 16}
\begin{minipage}{0.5\linewidth}
  \centering
  \includegraphics[width=0.95\linewidth]{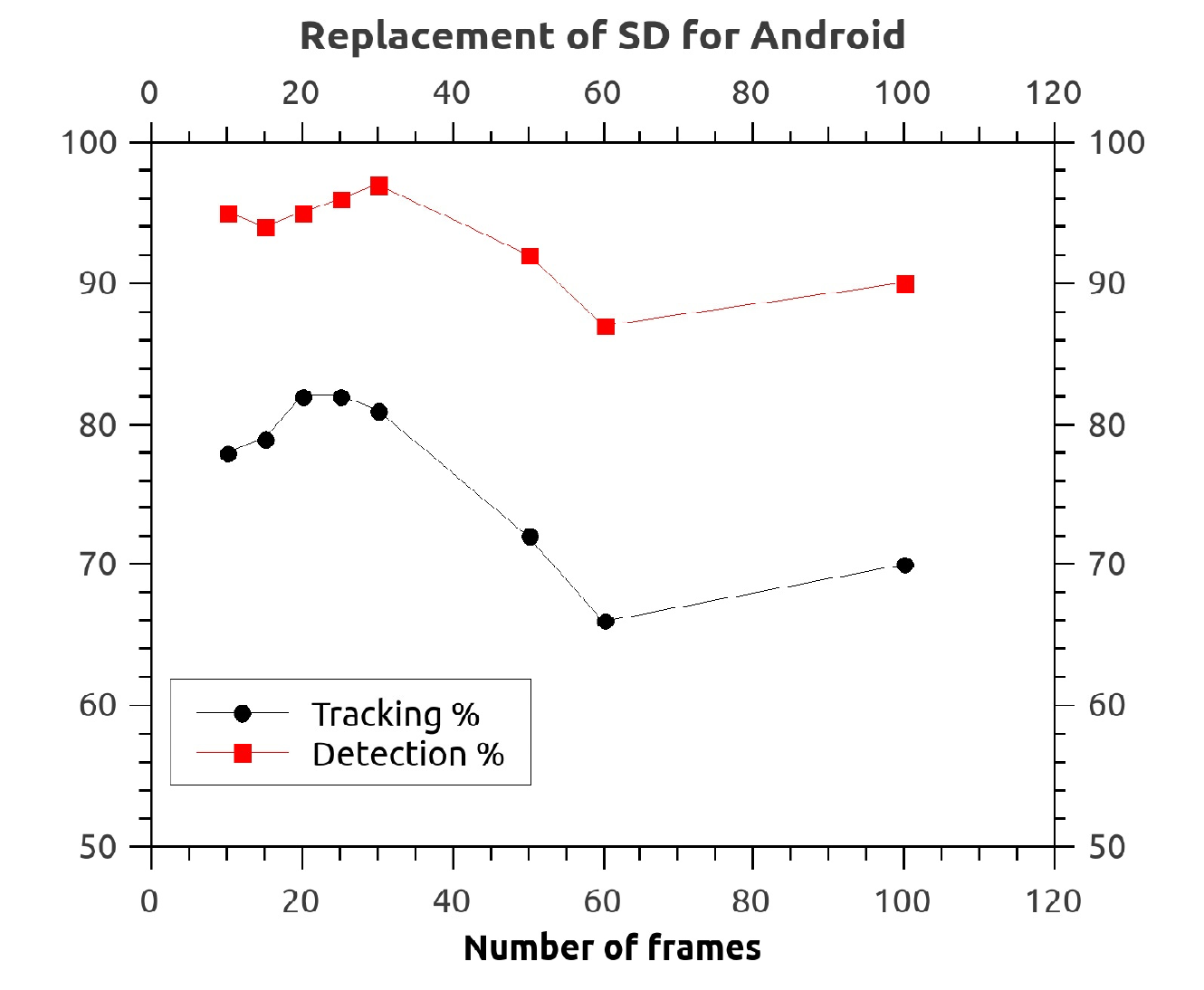}
  \label{frames}
\end{minipage}%
\begin{minipage}{0.5\linewidth}
  \centering
  \includegraphics[width=0.95\linewidth]{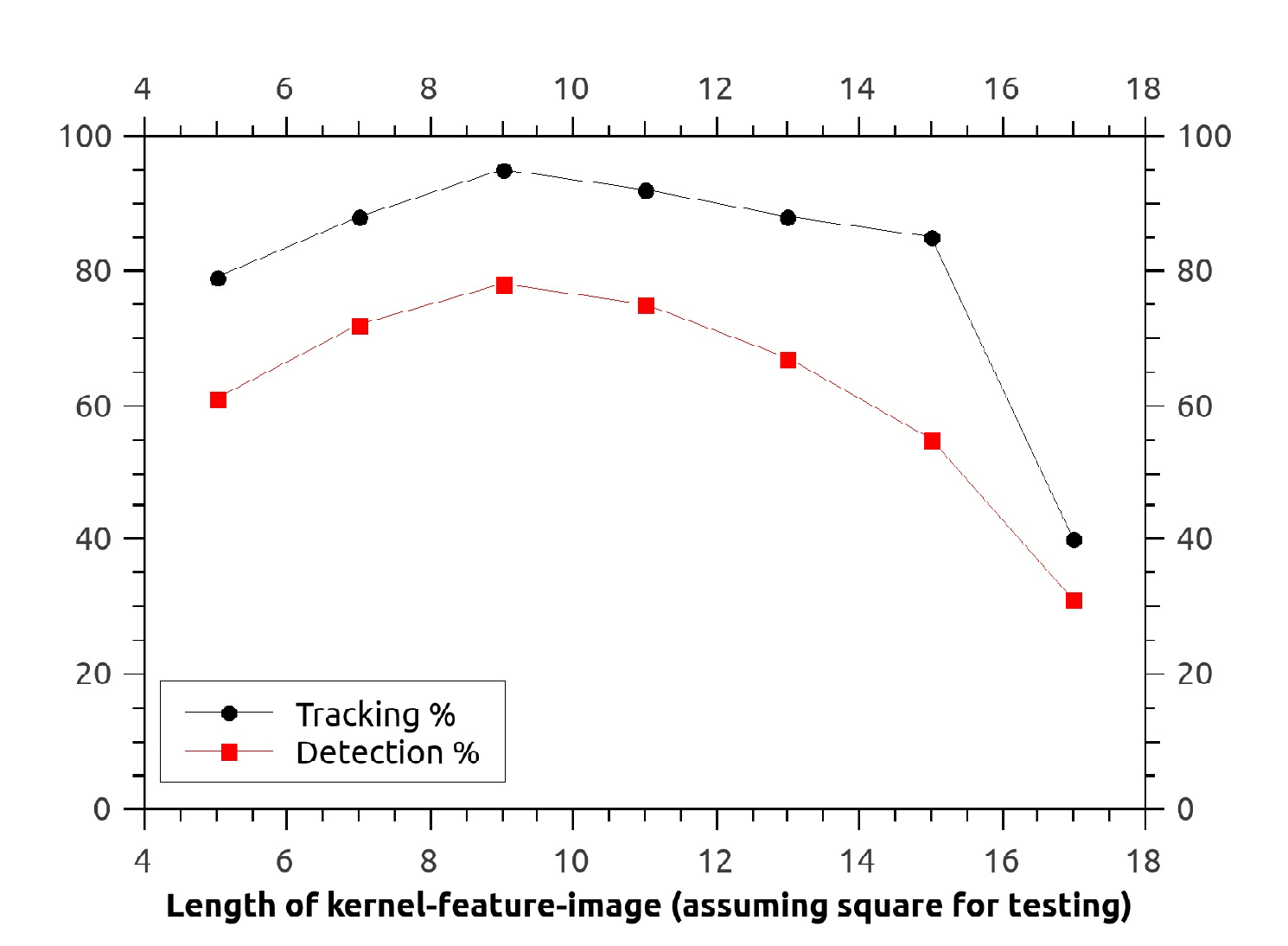}
  \label{imageKernel}
\end{minipage}
\captionof{figure}{Bottom-left: Replacement of SD and hard-coding number of frames of image-kernel to avoid Garbage collection (without rooting phone) and Bottom-right: Tracking for various image-feature-kernel sizes.}
\end{center}

\section{Discussion, Conclusion and future works}

We have thus built an autonomous follow-anything drone which can be used for 'fleet learning', reconnaissance and more. Our drone isn't just a selfie drone or follow-me drone although it is does that too without hassles. The core of our software code has been open-sourced under GPL licence (Creative-common's Attribution-NonCommercial 4.0 International isn't applicable to softwares) \cite{traquad35}. One of the additional challenges that we would definitely consider is oblique tracking at 45 degrees or any angle as mentioned in Unreal Engine simulation \cite{45degreeSimulator} as our drone can track only if camera is aligned horizontally at 0 degrees. Regarding simulations, many open-source applications run Noveuau driver (including NVIDIA) and we believe that an entire package consisting of graphic driver setup for Gazebo with some more custom models of Rover can improve research-capabilities of our simulator. Regarding Android, as of 21 January 2018, we have come across Kotlin. Kotlin is fully supported for Android and runs on its JVM with same garbage collection policies. A research area regarding a possibility of unified interface for Web-browser and android app is a possible resarch aspect (But, Kotlin coroutines like async/wait are in experimental design and neither Android nor Kotlin make guarantees. Also, Kotlin/Native is currently in development; preview releases are available. (\ref{Appendix 34}) ). As electronics continue to get cheaper, these drones can then be costing virtually nothing and advanced boards consisting of multi-core processors with multi-PRU units for high-end information-processing. Details regarding cost of our drone can be found in \hyperref[Appendix B]{Appendix B}. In the future, it may be possible that drones would be fitted with four wheels (may be a car-drone to save power; Initial stages are being experienced practically (\ref{Appendix 26})) and used as flying-robot-pets/assistant robots when chat-bots are integrated. This can also track suspicious objects and can thus be used in remote navigation and might even aid in the development of technological singularity. In the near future, passenger drones like Volocopter may be "tracked" (just like swarm formation of birds) and our object tracking algorithm would help in transfer machine learning for autonomous navigation of such passenger drones.(\ref{Appendix 32})
\\\\
We have demonstrated an image-processing based tracking drone; But, that is not the optimal one. The machine learning aspect can include contextual information (drones must not land and move ahead, cars do not fly etc) and include associative learning with machine knowledge graphs (cars are not supersonic generally etc) and tune the parameters using knowledge graphs\cite{learnAndThinkLikePeople}. An ensemble classifier can then be used across these classifiers may be with algorithms like TLD, non-linear classifier SVM and a minimalist-NEAT (if invented for softwares) and other frequency-transform based methods like SRDCF and non-ORB-FAST which may suffer from centroid-displacement and may contain relatively unneeded noise-reduction parameters\cite{trackingUAV}. We have considered object tracking for one drone. It may be possible that we would have multiple drones track the same object or there are mid-air collisions with other drones. For this, some planning software may have to be researched upon (In our draft version of 6 January 2017, we had mentioned flying pets and quad-copter fitted with wheels. We are adding a citation for that in revision as there is a planning algorthm for the same \cite{bothDriveAndFly}). However. the extent of the scope of such an extension of machine learning is still an open-question.

\section*{Acknowledgments}
We would like to thank Erle Robotics team for the prompt assistance in correction of documentation of their Copter's SITL and also Vladimir Ermakov, leading developer of MAVROS for assistance regarding MAVROS (although we could not integrate MAVROS into second instance of Ardupilot). We would like to Amrinder Singh Randhawa and Harsha H N, (then students) alumni of Electronics and Communication, M S Ramaiah Institute of Technology for sharing the expenses of the college project (about 100 dollars each) and providing valuable assistance and guidance while programming on WiFi hotspot. We would like to thank Venkat of Edall Systems, Bangalore for generous assistance in programming with APM PWM RC Override input. However, we would also disclose that these people were not directly involved in the development of this journal and there are no conflicts of interest (We have deliberately avoided funding for this journal to avoid conflicts of interest).



\appendix

\footnotesize
\sloppy
\section*{Appendix A: Further online resources and websites}

\subsection{AM335x U-Boot User's Guide, Texas Instruments.}
\label{Appendix 1}
\url{http://processors.wiki.ti.com/index.php/AM335x\_U-Boot\_User's\_Guide}

\subsection{ArduPilot Documentation.}
\label{Appendix 2}
a. Licence. \url{http://ardupilot.org/dev/docs/license-gplv3.html\#license-gplv3}
\newline b. Powering the APM2. \url{http://ardupilot.org/copter/docs/common-powering-the-apm2.html}
\newline c. APM 2.5 and 2.6 Overview. \url{http://ardupilot.org/copter/docs/common-apm25-and-26-overview.html#common-apm25-and-26-overview}
\newline d. Starting up and calibrating Plane. (similar to Copter) \url{http://ardupilot.org/plane/docs/starting-up-and-calibrating-
arduplane.html}
\newline e. Traditional Helicopter – Connecting the APM. (similar to Copter) \url{http://ardupilot.org/copter/docs/traditional-helicopter-
connecting-apm.html}
\newline f. Pre-Arm Safety Check. \url{http://ardupilot.org/copter/docs/prearm\_safety\_check.html}
\newline g. Accelerometer Calibration in Mission Planner. \url{http://ardupilot.org/copter/docs/common-accelerometer-calibration.html}
\newline h. Optional Hardware. \url{http://ardupilot.org/copter/docs/common-optional-hardware.html}
\newline i. Threading. \url{http://ardupilot.org/dev/docs/learning-ardupilot-threading.html}
\newline j. Command Line Interface to Configure Copter. \url{http://ardupilot.org/dev/docs/using-the-command-line-interpreter-to-
configure-apmcopter.html}
\newline k. Battery Fail-safe. \url{http://ardupilot.org/copter/docs/failsafe-battery.html}
\newline l. Loiter Mode. \url{http://ardupilot.org/copter/docs/loiter-mode.html}
\newline m. 3DR Power Module. \url{http://ardupilot.org/copter/docs/common-3dr-power-module.html}
\newline n. UAVCAN ESCs. \url{http://ardupilot.org/copter/docs/common-uavcan-escs.html}

\subsection{Camera Capes of Beaglebone Black.}
\label{Appendix 3}
a. CircuitCo 3.1MP Cape, Element 14. \url{http://in.element14.com/circuitco/bb-bone-cam3-01/board-beaglebone-camera-3-1mp/dp/2144194}
\newline b. HD Camera Cape for BeagleBone Black, Texas Instruments.
\url{http://www.ti.com/devnet/docs/catalog/endequipmentproductfolder.tsp?actionPerformed=productFolder&productId=19580}

\subsection{Other open-source cameras.}
\label{Appendix 4}
a. PixyCam. \url{https://www.kickstarter.com/projects/254449872/pixy-cmucam5-a-fast-easy-to-use-vision-sensor/description}
\newline b. OpenMV. \url{https://www.kickstarter.com/projects/botthoughts/openmv-cam-embedded-machine-vision/description}
\newline c. Datasheet of OV2640 by UCtronics, distributor of ArduCam.
\url{http://www.uctronics.com/download/cam_module/OV2640DS.pdf}

\subsection{Beaglebone Black's WiFi-Direct Cape.}
\label{Appendix 5}
\url{http://boardzoo.com/index.php/catalog/product/view/id/146/category/8#.V9AckTXD71y}
\newline x. Optional non-standard software-defined-radio. \url{https://www.kickstarter.com/projects/mossmann/hackrf-an-open-source-
sdr-platform/description}

\subsection{WiFi vs Bluetooth comparison.}
\label{Appendix 6}
a. “How far does a WiFi-Direct Connection Travel?”, WiFi Alliance FAQs. \url{http://www.wi-fi.org/knowledge-center/faq/how-far-does-a-wi-fi-direct-connection-travel}
\newline b. Ian Paul, PCWorld, “WiFi-Direct vs Bluetooth 4.0: A battle for supremacy”
\url{http://www.pcworld.com/article/208778/Wi\_Fi\_Direct\_vs\_Bluetooth\_4\_0\_A\_Battle\_for\_Supremacy.html}
\newline c. Wi-Fi Peer-to-Peer. \url{https://developer.android.com/guide/topics/connectivity/wifip2p.html}

\subsection{Blender software.}
\label{Appendix 7}
\url{https://www.blender.org}

\subsection{Matlab Embedded Coder.}
\label{Appendix 8}
a. Functions and Objects Supported for C/C++ Code Generation – Alphabetical List, Mathworks documentation, retieved
using Internet Archive WaybackMachine, 8th March 2015.
\url{http://web.archive.org/web/20150308152443/http://in.mathworks.com/help/simulink/ug/functions-supported-for-code-generation--alphabetical-list.html}
\newline b. Coder.ceval, Mathworks documentation. \url{http://in.mathworks.com/help/simulink/slref/coder.ceval.html}
\newline c. Coder.extrinsic, Mathworks documentation. \url{http://in.mathworks.com/help/simulink/slref/coder.extrinsic.html}

\subsection{eCalc}
\label{Appendix 9}
\url{http://www.ecalc.ch/xcoptercalc.php}

\subsection{Dronecode's support for APM.}
\label{Appendix 10}
a. Initial core projects, FAQs, Dronecode. \url{https://www.dronecode.org/news-faq/faq}
\newline b. Linux Foundation and Leading Technology Companies Launch Open Source Dronecode Project, Linux Foundation
Newsletter, October 12 th 2014. \url{https://www.linuxfoundation.org/news-media/announcements/2014/10/linux-foundation-and-leading-technology-companies-launch-open}

\subsection{OpenCV's applications.}
\label{Appendix 11}
a. Stanley, Stanford's autonomous car. \url{http://www.willowgarage.com/pages/software/opencv}
\newline b. Falkor systems's open-source softwares. \url{https://github.com/FalkorSystems/}
\newline c. Moksha-4, M S Ramaiah Institute of Technology, IGVC 2014. \url{www.igvc.org/design/2014/12.pdf}

\subsection{Falkor system's software and business.}
\label{Appendix 12}
a. Why I Heart Engineering Shut Down, Technical.ly, October 27, 2014. \url{http://technical.ly/brooklyn/2014/10/27/i-heart-engineering-shut-down/}
\newline b. Falkor Systems's logo. \url{https://github.com/FalkorSystems/falkor_ardrone/tree/master/images}
\newline c. Usage of Haar and LBP classifiers. \url{https://github.com/FalkorSystems/falkor_ardrone/tree/master/cascade}
\newline d. PID code for AR Parrot drone. \url{https://github.com/FalkorSystems/falkor_ardrone/blob/master/nodes/ardrone_follow.py}
\newline e. Object-detection and tracking code with specific parameters.
\url{https://github.com/FalkorSystems/falkor_ardrone/blob/master/nodes/tracker.py}
\newline f. SURF with brute-force matching with specific parameters meant for Falkor logo.
\url{https://github.com/FalkorSystems/ardrone_autonomy_legacy/blob/master/src/ardrone_tracker.cpp}

\subsection{Percepto 999\$ kit.}
\label{Appendix 13}
\url{http://www.percepto.co/}

\subsection{FlightGear software flight simulator.}
\label{Appendix 14}
a. Setting up SITL on Linux. \url{http://ardupilot.org/dev/docs/setting-up-sitl-on-linux.html}
\newline b. Documentation. \url{http://www.flightgear.org/docs.html}

\subsection{Arducopter parameters of APM.}
\label{Appendix 15}
\url{http://ardupilot.org/copter/docs/parameters.html#parameters}

\subsection{PWM information.}
\label{Appendix 16}
a. PWM Servos and Motor Controllers. \url{https://pixhawk.org/users/actuators/pwm_escs_and_servos}
\newline b. RC Transmitter Flight Mode Configuration. \url{http://ardupilot.org/copter/docs/common-rc-transmitter-flight-mode-
configuration.html}
\newline c. Auxiliary Function Switches. \url{http://ardupilot.org/copter/docs/channel-7-and-8-options.html}

\subsection{ROS and Gazebo installation.}
\label{Appendix 17}
a. Supported operating systems for ROS. \url{http://wiki.ros.org/ROS/Installation}
\newline b. Supported operating systems for Gazebo. \url{http://gazebosim.org/tutorials?cat=install}
\newline c. Common operating system for Gazebo and ROS. \url{http://gazebosim.org/tutorials?tut=ros_installing&cat=connect_ros}

\subsection{USB EEPROM Issues.}
\label{Appendix 18}
a. USB driver installation warning.
\url{http://elinux.org/Beagleboard:BeagleBone#Trouble_Installing_USB_Drivers_.5BA4_and_Earlier.5D}
\newline b. AN\_136\_Hi-Speed Mini Module EEPROM Disaster Recovery, FT\_000209, Version 1.0, Clearance No. 138.
\url{http://www.ftdichip.com/Support/Documents/AppNotes/AN_136\%20Hi\%20Speed\%20Mini\%20Module\%20EEPROM\%20Disaster\%20Recovery.pdf}
\newline c. CAT24C256 CMOS Serial EEPROM's datasheet. \url{https://cdn.sparkfun.com/datasheets/Dev/Beagle/CAT24C256-
D.PDF}
\newline d. Software Application Development, D2XX Programmer's Guide, Future Technology Devices International ltd, version
1.3, 2012-02-23.
\url{http://www.ftdichip.com/Support/Documents/ProgramGuides/D2XX_Programmer's_Guide%28FT_000071%29.pdf}
\newline e. D2XX drivers. \url{http://www.ftdichip.com/Drivers/D2XX.htm}
\newline f. Adafruit FT232H Breakout Board. \url{https://learn.adafruit.com/adafruit-ft232h-breakout/mpsse-setup}

\subsection{GNU Octave software.}
\label{Appendix 19}
\url{https://www.gnu.org/software/octave/}

\subsection{Standard comparison of embedded-systems.}
\label{Appendix 20}
a. Parallela board's specifications. \url{https://www.parallella.org/board/}
\newline b. “Creating a \$99 parallel computing machine is just as hard as it sounds”, Jon Brodkin, Ars Technica, 30 th July 2013.
\url{http://arstechnica.com/information-technology/2013/07/creating-a-99-parallel-computing-machine-is-just-as-hard-as-it-sounds/}
\newline c. Embedded Linux Board Comparison, Tony DiCola, 26 th September 2014. \url{https://cdn-learn.adafruit.com/downloads/pdf/embedded-linux-board-comparison.pdf}
\newline d. Comparisons of Embedded Boards, LCD Displays and Sensors, Douglas McIlrath.
\url{http://people.csail.mit.edu/dcurtis/assistive_devices_for_healthcare/Comparisons.html}
\newline e. “A Comparison of FPGA and GPU for Real-time Phased-based Optical-flow, Stereo and Local Image Features”, Karl
Pauwels, Matteo Tomasi, Javier Diaz, Eduardo Ros and Marc M Van Hulle, IEEE Transactions on Computers, Vol 61,
No 7, pages 999- 1012, July 2012. \url{http://ieeexplore.ieee.org/document/5936059/}
\newline f. MyHDL software-tool – Python to VHDL conversion tool. \url{http://www.myhdl.org/}

\subsection{ROS Gazebo SITL.}
\label{Appendix 21}
a. Using ROS/Gazebo simuator with SITL, Ardupilot documentation. \url{http://ardupilot.org/dev/docs/using-rosgazebo-
simulator-with-sitl.html}
\newline b. Mavros, ROS package, Erle-gitbook. \url{http://erlerobot.github.io/erle_gitbook/en/mavlink/ros/mavros.html}
\newline c. Ros2 alpha 7 July 2016 (latest) release. \url{https://github.com/ros2/ros2/wiki/Alpha7-Overview}
\url{http://design.ros2.org/articles/why_ros2.html}
\newline d. Erle-Copter and Erle-Rover in SITL software-simulation. \url{http://erlerobotics.com/docs/Simulation/index.html}
\newline e. SITL simulator, Ardupilot documentation. \url{http://ardupilot.org/dev/docs/sitl-simulator-software-in-the-loop.html#sitl-simulator-software-in-the-loop}
\newline f. APM PID Autotune issues, Ardupilot documentation. \url{http://ardupilot.org/copter/docs/autotune.html}

\subsection{“WiFi and Bluetooth - Interference issues”, HP computers, January 2002.}
\label{Appendix 22}
\url{http://www.hp.com/sbso/wireless/images/WiFiBlue.pdf}

\subsection{Drone stores in Bangalore.}
\label{Appendix 23}
a. RCbazaar \url{http://www.rcbazaar.com/default.aspx}
\newline b. Edall Stores \url{http://www.edallhobby.com/en/}

\subsection{Altitude hold mode, Ardupilot documentation.}
\label{Appendix 24}
\url{http://ardupilot.org/copter/docs/altholdmode.html}

\subsection{"Behind The Crash Of 3D Robotics, North America's Most Promising Drone Company", Forbes, October 5th 2016.}
\label{Appendix 25}
\url{www.forbes.com/sites/ryanmac/2016/10/05/3d-robotics-solo-crash-chris-anderson}

\subsection{"First passenger drone makes its debut at CES 2016", The Guardian, 7th January 2016.}
\label{Appendix 26}
\url{https://www.theguardian.com/technology/2016/jan/07/first-passenger-drone-makes-world-debut}

\subsection{Setup of Pioneer 3DX in ROS Gazebo SITL}
\label{Appendix 27}
a. "ROS indigo and Gazebo2 Interface for the Pioneer3dx Simulation Ubuntu 14.04 LTS (Trusty Tahr)", Jen Jen Chung, February 22nd 2016.
\url{http://people.oregonstate.edu/~chungje/Code/Pioneer3dx\%20simulation/ros-indigo-gazebo2-pioneer.pdf}
\newline b. ROS Wiki. \url{http://wiki.ros.org/action/show/Robots/AMR\_Pioneer\_Compatible}

\subsection{Flaws of our ROS Gazebo SITL}
\label{Appendix 28}
a. Segmentation fault during start-up. \url{http://wiki.ros.org/rviz/Troubleshooting#Segfault_during_startup}
\newline b. Noveau graphic driver issue if installed incorrectly.
\url{http://sdk.rethinkrobotics.com/wiki/Gazebo_Troubleshooting}

\subsection{Intel EUCLID and UP boards}
\label{Appendix 29}
a. Intel EUCLID.
\url{https://www.theverge.com/circuitbreaker/2017/5/23/15682172/intel-euclid-robotics-development-kit-launch-date-price}
\newline
b. UP Squared.
\url{https://www.kickstarter.com/projects/802007522/up-squared-the-first-maker-board-with-intel-apollo/updates}
\newline
c. UP board.
\url{https://www.kickstarter.com/projects/802007522/up-intel-x5-z8300-board-in-a-raspberry-pi2-form-fa}
\newline
d. UP Core.
\url{https://www.kickstarter.com/projects/802007522/up-core-the-smallest-quadcore-x86-single-board-com/description}

\subsection{Grove connectors and ARM Microcontroller profiles}
\label{Appendix 30}
a. Grove connectors for Seeed Studio's Beaglebone Green.
\url{https://www.seeedstudio.com/category/Grove-c-45.html}
b. ARM Microcontroller profiles
\url{http://infocenter.arm.com/help/index.jsp?topic=/com.arm.doc.dui0471i/BCFDFFGA.html}

\subsection{JeVois: Open Source camera}
\label{Appendix 31}
\url{https://www.kickstarter.com/projects/1602548140/jevois-open-source-quad-core-smart-machine-vision}

\subsection{Volocopter: Passenger drone takes flight for the first time in US, January 8 2018}
\label{Appendix 32}
\url{https://www.theverge.com/transportation/2018/1/8/16866662/volocopter-flying-taxi-first-us-flight-intel-ces-2018}

\subsection{GroPro and Lily drones' shutdown}
\label{Appendix 33}
a. Lily's shut down: \url{http://www.bbc.com/news/technology-38595473}\\
b. GoPro's shutdown:
\url{https://techcrunch.com/2018/01/09/gopro-ceo-explains-shutdown-of-drone-program/}\\

\subsection{Kotlin on Android and native compilation}
\label{Appendix 34}
a. Kotlin/Native is currently in development; preview releases are available.
\url{https://kotlinlang.org/docs/reference/native-overview.html}\\
b. Kotlin's coroutines and Garbage collection.
\url{https://developer.android.com/kotlin/faq.html}

\section*{Appendix B: Cost of TraQuad and comparison}
\label{Appendix B}
\begin{table}
\label{Table 3}
  \centering
  \caption{Cost of TraQuad; Some were purchased online, others at RCbazaar (\ref{Appendix 23}a) and Edall stores (\ref{Appendix 23}b).}
  \resizebox{\linewidth}{!}{
    \begin{tabular}{cccc}
    \textbf{Hardware Components} & \textbf{Price (Rs)} & \textbf{Quantity} & \textbf{Total price (Rs)} \\
    Avionic C2830 KV850 QUAD brushless motor & 1090  & 4     & 4360 \\
    Ardupilot Mega 2.6 Flight Controller Arduino Compatible & 3470  & 1     & 3470 \\
    DYS Electronic Speed Controller(ESC) 30A & 460   & 4     & 1840 \\
    Hiller Chassis Q450 - PCB version (kit) & 1310  & 1     & 1310 \\
    Wolfpack 2200mAh 25C 11.1V Battery & 1026  & 1     & 1026 \\
    APM 2.6 Power Module (5.3V) & 650   & 1     & 650 \\
    Battery Charger (local brand) & 600   & 1     & 600 \\
    Miscellaneous & 600   & 1     & 600 \\
    Arduino UNO & 520   & 1     & 520 \\
    Bluetooth (HCSR04) & 500   & 1     & 500 \\
    Landing Gear & 488   & 1     & 488 \\
    10"x4.5" propeller-set & 471   & 1     & 471 \\
    HCSR04 ultrasonic sensor & 210   & 2     & 420 \\
    Voltage buzzer & 300   & 1     & 300 \\
    Mobile phone holder & 148   & 1     & 148 \\
    Bullet connector and Servo-lead set & 274   & 1     & 274 \\
    \textbf{Grand Total} & \textbf{} & \textbf{} & \textbf{16977} \\
    \end{tabular}}
  \label{tab:traquadCost}%
\end{table}%

Thus, we were able to maintain low-cost of the quadcopter by beginning with the electronics and working on the mechanical aspects later. This cost is significantly lesser than the normal FPV or GPS based or device's Wifi/Bluetooth-triangulation based follow-me drones. (Note: Solo and Parrot are exceptions as they offer developer options in the softwares although they are FPV drones when unboxed) The cost comparison has been included in \hyperref[Table 4]{Table 4}. Break-down analyses of cost of TraQuad has been included in \hyperref[Table 3]{Table 3}.

\begin{table*}[htbp]

\label{Table 4}
\resizebox{\textwidth}{!}{
\begin{tabular}{|l|l|l|l|l|}
\hline
\textbf{Name} & \textbf{Principle of Operation} & \textbf{Marketing stage} & \textbf{Cost} & \textbf{Webiste link}                               \\\hline
TraQuad       & Imaging based (follow anything) & -               & 17000 Rs                              & www.github.com/traquad                     \\\hline
AirDog        & \shortstack{Hardware based (Bluetooth)\\follow-me drone}                & Existing        & 1600 \$ = 107312 Rs & www.airdog.com                     \\\hline
Solo          & FPV drone (but customisable) & Shut down        & 800 \$ = 53656 Rs                     & store.3dr.com/products/solo                \\\hline
DJI           & \shortstack{FPV drone (proprietary),\\GPS and proprietary vision based follow-me}    & Existing & 500$ to 6000$ = 37420 Rs to 449000 Rs & store.dji.com/                             \\\hline
Xiaomi        & FPV drone (proprietary) & Existing & 519\$ = 34812 Rs & www.xiaomidevice.com/xiaomi-drone.html     \\\hline
Parrot        & FPV drone (but customisable) & Existing & 250 to 700 pounds = 18700 Rs to 52400 Rs & www.parrot.com/us/Drones                   \\\hline
Nixie         & No description -- "Nixie is coming soon" & Pre-order & Yet to hit market & www.flynixie.com/                          \\\hline
Lily          & \shortstack{Uses tracking device (GPS based) \\follow-me human tracker drone} & Shut down (\ref{Appendix 33}) & 920 \$ = 61704 Rs & www.lily.camera                     \\\hline
Zero Zero Robotics & Proprietary selfie-drone & Existing & \shortstack{Yet to hit market in some countries like\\India (as of 21 January 2018)} & www.gethover.com \\\hline
GoPro         & FPV drone & Shut down (\ref{Appendix 33}) & 800\$ = 59872 Rs & shop.gopro.com/International/karma \\\hline
AirSelfie     & Selfie only drone & Pre-order & 360\$ = 27000 Rs & www.kickstarter.com/projects/1733117980/airselfie \\\hline
Intel Falcon 8/8+ & Proprietary "Circle around" mode & Existing & Only for businesses & \shortstack{www.intel.com/content/www/us\\/en/products/drones/falcon-8.html} \\\hline
\shortstack{MAVinci hand-launched\\ SIRIUS aeroplane} & & & & \\\hline
\end{tabular}
}
\begin{footnotesize}
\newline
Note 1: 1\$ = 67.07 Rs, 1 pound = 74.04 Rs\\
Note 2: We have ignored some of the plane versions which are meant only for businesses like MAVinci Sirius hand-launched aeroplanes and other non-standard options.
\end{footnotesize}
\caption{Cost based comparison}
\end{table*}

\begin{figure}
\label{Figure 18}
\centering
\includegraphics[width=\linewidth]{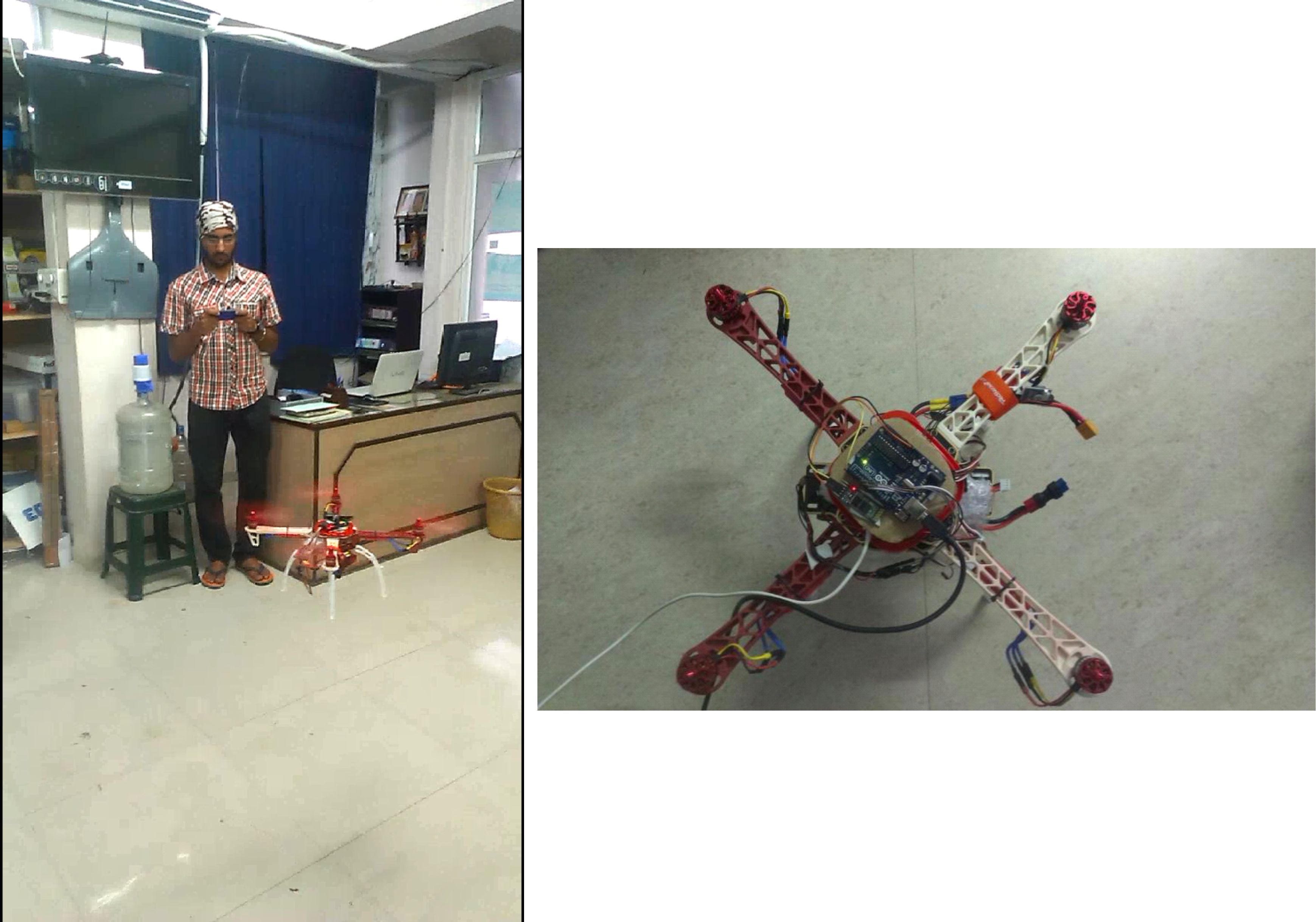}
\caption{TraQuad's mechanical design getting tested in the left; Electronics is being shown in the right.}
\end{figure}
\newpage
\section*{Appendix C: Resultant statistical distribution function}
\label{Appendix C}

$ce^{-\frac{x^2}{2\sigma^2}}$ $\implies$ Normalised probability of form $ce^{-\frac{x^2}{2\sigma^2}}$ having a total area of the form $c\int_{-\infty}^{\infty} e^{-\frac{x^2}{2\sigma^2}} dx = 1$.

$\sigma\sqrt{2}c\int_{0}^{\infty}e^{-u^2} du = 1$ where $u = \frac{x}{\sigma\sqrt{2}}$ $\implies$ $2\sigma^{2}c^{2}\int_{0}^{\infty}\int_{0}^{\infty}e^{-(v^{2} + w^{2})} dv dw = 1$ $\forall$ \textit{v,w} $\equiv$ \textit{u}.\\

In polar form, $2\sigma^{2}c^{2}\int_{0}^{2\pi}\int_{0}^{\infty}e^{- r^{2}} rdr  d\theta= 1$ $\because$ $Area_{cartesian} = Area_{polar}$ \\ $\because dArc \times dr = du \times dw \implies dvdw = (rd \theta)(dr)$.

$-\sigma^{2}c^{2}\int_{0}^{2\pi}\int_{0}^{\infty}e^{-m^{2}} dm d\theta = 1$ $\forall$ $m \equiv u$ $\implies$ $c = \frac{1}{\sigma\sqrt{2\pi}}$

When mean is non-zero, the resultant is a Gaussian function which we used for two-dimensions.

$z = f(x,y) = \frac{1}{\sigma\sqrt{2\pi}}e^{-\frac{(x-\mu_x)^2 + (y-\mu_y)^2}{2\sigma^2}}$


\end{document}